%% file: neurips_2026.tex
\documentclass{article}

\PassOptionsToPackage{numbers, compress}{natbib}
\usepackage{booktabs,tabularx, colortbl} 
\usepackage{multirow}
\usepackage{float}
\usepackage{amsfonts} 
\usepackage{bm}
\usepackage{mathtools}
\usepackage{amsmath}
\usepackage{algorithm}
\usepackage{algorithmic}
\usepackage{subcaption}

\usepackage{amssymb}
\usepackage{amsthm}

\usepackage{wrapfig}
\usepackage{booktabs}
\usepackage{multirow}
\usepackage{graphicx}
\usepackage{colortbl}
\usepackage{array}
\usepackage{enumitem}

\definecolor{lightred}{RGB}{255, 230, 230}
\definecolor{lightgreen}{RGB}{230, 255, 230}

\usepackage[most]{tcolorbox}

\usepackage{tabularx}
\usepackage{colortbl} 
\definecolor{s-gray}{gray}{0.95} 

\definecolor{ggreen}{rgb}{0.0, 0.6, 0.0}
\definecolor{rred}{rgb}{0.75, 0.0, 0.0}
\definecolor{bblue}{rgb}{0.13, 0.67, 0.8}

\definecolor{BoxBackground}{RGB}{240, 240, 240} 
\definecolor{BoxFrame}{RGB}{0, 0, 0} 
\definecolor{TitleBackground}{RGB}{0, 0, 0} 
\definecolor{TitleText}{RGB}{255, 255, 255} 
\definecolor{deepgreen}{RGB}{0,100,0}
\tcbset{
  academicbox/.style={
    boxsep=5pt,
    left=2pt,
    right=2pt,
    bottom=0.5pt,
    boxrule=0.5pt,
    colback=BoxBackground,
    colframe=BoxFrame,
    colbacktitle=TitleBackground,
    coltitle=TitleText,
    enhanced,
    attach boxed title to top left={yshift=-0.1in,xshift=0.1in},
    boxed title style={boxrule=0pt,colframe=white},
    title={#1},
  }
}

\newtcolorbox{AcademicBox}[1][]{academicbox=#1}
\definecolor{SoftBlue}{RGB}{135, 206, 250}  
\definecolor{SoftOrange}{RGB}{255, 224, 178} 
\definecolor{SoftGreen}{RGB}{144, 238, 144}  
\definecolor{CorrectGreen}{RGB}{76, 175, 80} 
\definecolor{ErrorRed}{RGB}{211, 47, 47} 

\usepackage{nicefrac}

 \usepackage[preprint]{neurips_2026}

\usepackage[utf8]{inputenc} 
\usepackage[T1]{fontenc}    
\usepackage{hyperref}       
\usepackage{url}            
\usepackage{booktabs}       
\usepackage{amsfonts}       
\usepackage{nicefrac}       
\usepackage{microtype}      
\usepackage{xcolor}         
\usepackage{cleveref}
\title{Mechanistic Insights into Functional Sparsity in Multimodal LLMs via CoRe Heads}

\author{
Ruoxi Sun$^{1}$,
Quantong Qiu$^{1}$,
Juntao Li$^{1}$\thanks{Corresponding author.},
Zecheng Tang$^{1}$,
Yihang Lou$^{2}$,
Min Zhang$^{1}$ \\
$^{1}$ Soochow University \\
$^{2}$ Peking University \\
\texttt{\{ljt\}@suda.edu.cn} \\
}

\begin{document}

\maketitle

\vspace{-1.5em}
\input{paper/0-abs}
\input{paper/1-intro}
\input{paper/related_work}

\input{paper/3-method}
\input{paper/4-exp}

\input{paper/5-analysis}

\input{paper/6-conclusion}

\small

\bibliographystyle{plain}
\bibliography{main}

\clearpage


\appendix

\input{paper/7-appendix}

\end{document}

%% file: paper/0-abs.tex
\begin{abstract}
While Multimodal Large Language Models (MLLMs) demonstrate remarkable proficiency on complex vision-language tasks, the mechanisms by which they extract query-relevant visual features from complex, noisy contexts remain opaque.
In this paper, we present an in-depth interpretability study that uncovers a profound structural property within MLLMs: \textbf{\textit{functional sparsity}} in cross-modal retrieval. Leveraging a token-level metric termed Retrieval Attention Mass (RAM), we identify and characterize a highly specialized subset of attention heads, referred to as \textbf{\textit{Context-aware Retrieval (CoRe) heads}}.
Across diverse visual domains and model scales, we observe a clear functional division: CoRe heads act as dedicated information extractors, while most other heads distribute attention over broader contextual regions. 
Causal interventions further demonstrate the necessity of these specialized heads. Ablating only the top 5\% of CoRe heads causes significant degradation in multimodal reasoning performance, whereas ablating lower-ranked heads has minimal effect. 
Moreover, acceleration experiments validate the utility of CoRe heads, showing that leveraging this localized sparsity significantly accelerates inference while maintaining robust task performance.
Our findings reveal a structural principle of functional sparsity within MLLMs, refining the current understanding of mechanistic interpretability and laying a theoretical foundation that can inspire future architecture design and model optimization.

\end{abstract}


%% file: paper/1-intro.tex
\section{Introduction}
\label{sec:intro}

\begin{wrapfigure}{r}{0.5\linewidth}
\centering
\vspace{-1.5em}
\includegraphics[width=\linewidth]{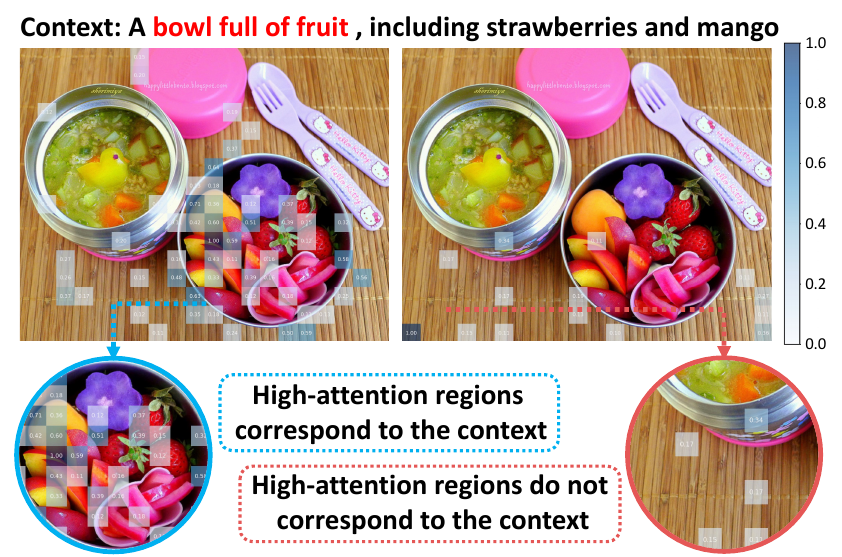}
\caption{Functional specialization in MLLM attention heads on RefCOCOg.
\textbf{Left (CoRe Heads):} High-attention regions correspond to context-relevant objects. 
\textbf{Right (Bottom Heads):} High-attention regions show week context-relevant.}
\label{fig:refcocog}
\vspace{-2.8em}
\end{wrapfigure}


Multimodal Large Language Models (MLLMs) have demonstrated remarkable capabilities in complex vision-language tasks~\citep{li2023blip, tang2025video, yin2024survey}. 
These models map high-dimensional visual signals into the semantic space of large language models~\citep{liu2023visual}.
However, real-world visual inputs are often cluttered with redundant information. 
Consequently, robust multimodal reasoning is dependent on the model's capacity to selectively retrieve and isolate sparse, task-relevant visual cues from these cluttered or complex spatiotemporal scenes.
Yet, despite the profound empirical success of MLLMs, the precise internal mechanisms governing this critical cross-modal information retrieval remain unexplored, presenting a significant gap in our understanding of multimodal mechanistic interpretability.

\begin{figure}[t]
    \centering
    \includegraphics[width=\linewidth]{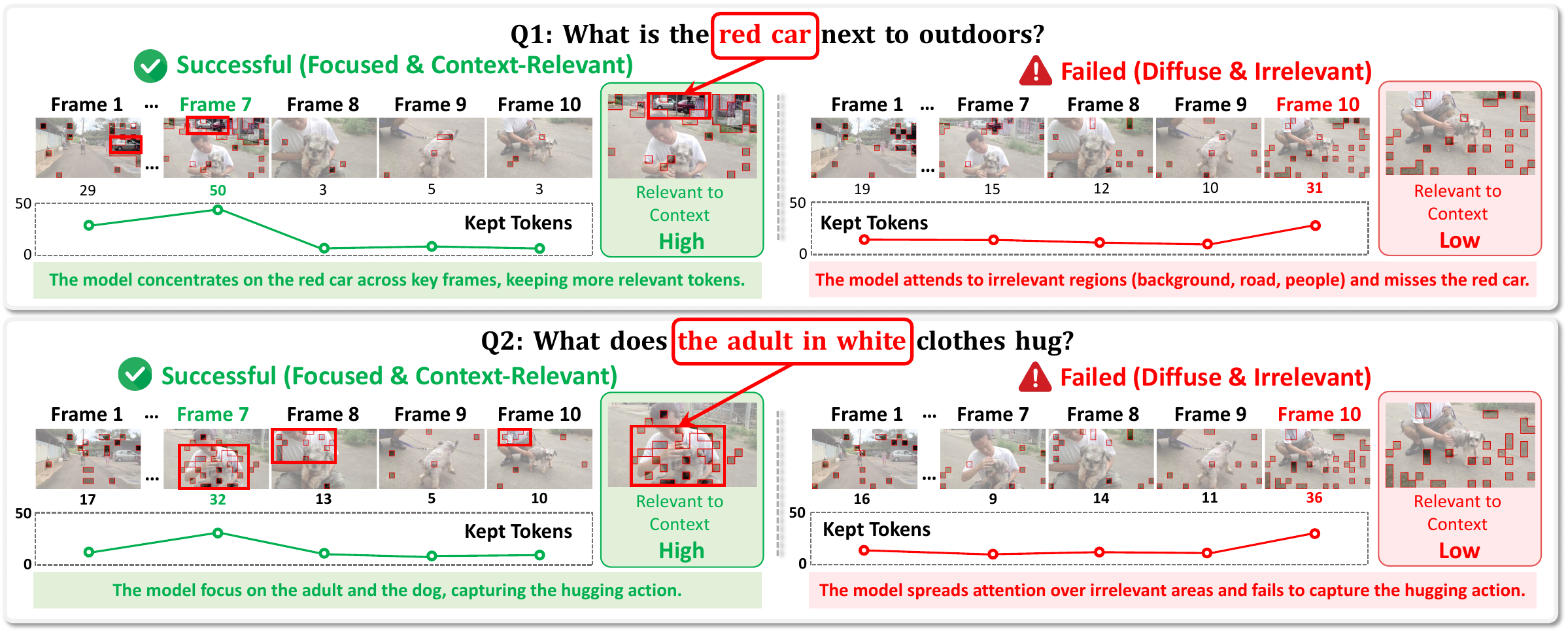}
    \caption{Mechanistic evidence of functional specialization in MLLMs attention heads on VidSTG. 
    \textbf{Left (CoRe Heads):} A sparse subset of specialized heads acts as precise information extractors, surgically isolating context-relevant entities (e.g., ``red car'', ``adult in white'') by filtering background noise across key frames. 
    \textbf{Right (Bottom Heads):} The vast majority of heads exhibit semantic dispersion, scattering attention across irrelevant regions and failing to ground the instruction.}
    \vspace{-1em}
\label{fig:attention_visualization}
\end{figure}

While recent interpretability studies have identified specialized attention heads in MLLMs~\citep{bi2025unveiling, ma2026attention, basile2025head}, how these models perform cross-modal information retrieval remains poorly understood. 
Current frameworks rely on coarse statistical heuristics such as spatial entropy~\citep{ma2026dear} and attention aggregation~\citep{li2026causal}, typically evaluated on simple static images. 
Consequently, these approaches struggle to capture fine-grained, query-conditioned token-level attention in dense or complex spatiotemporal scenes. 
Without rigorous quantitative methods to characterize these retrieval mechanisms, understanding MLLM failure modes and internal efficiency remains limited.

We find that within MLLMs, \textit{a specialized subset of attention heads consistently captures task-relevant visual information during multimodal understanding, which we called CoRe head. }
To identify these heads, we quantify the attention mass from context tokens to semantically relevant visual tokens.
We evaluate this across multiple complementary datasets covering different visual domains, including video-based spatio-temporal reasoning (VidSTG~\citep{zhang2020does}), document layout understanding (MMDocIR~\citep{dong2025mmdocir}), object-level visual grounding (RefCOCOg~\citep{kazemzadeh-etal-2014-referitgame}), and long-context multimodal reasoning (MMLongBench~\citep{wang2025mmlongbenchbenchmarkinglongcontextvisionlanguage}).
We further analyze a range of MLLMs, including Llava-onevision~\citep{li2025llavaonevision}, InternVL3.5~\citep{wang2025internvl3_5}, and the Qwen3-VL family~\citep{Qwen3-VL}, across different parameter scales.

Our analysis reveals a distinct functional dichotomy where CoRe heads execute localized visual extraction while standard vision heads facilitate global feature aggregation.
Furthermore, we note that as model capacity scales, these CoRe heads progressively concentrate in middle-to-late layers.
Evaluations across diverse multimodal datasets reveal a globally consistent yet locally variant paradigm: approximately 30 specific heads remain universally activated, while others dynamically adapt to specific data distributions. 
To rigorously ascertain their functional necessity, we conduct causal interventions via attention masking. 
Crucially, ablating just the top 5\% of CoRe heads causes a significant performance drop, confirming their indispensable role in cross-modal visual retrieval.

Spatial visualizations of attention weights further reveal a stark functional dichotomy during vision-language reasoning. 
As shown in Figures~\ref {fig:refcocog} and ~\ref {fig:attention_visualization}, across both image and video modalities, the top-ranked CoRe heads precisely localize context-relevant visual entities. 
In contrast, lower-ranked heads exhibit diffuse activation patterns, predominantly covering non-salient background regions. Corroborating our quantitative ablations, these observations confirm that fine-grained, context-aware visual selection is exclusively governed by a sparse subset of highly specialized heads.

The discovery of CoRe heads offers profound implications for the mechanistic interpretability and optimization of MLLMs.
Our findings reveal a structured organization in multimodal processing, where a sparse, consistent subset of heads governs cross-modal extraction across diverse modalities and scales.
As evidenced by our causal interventions, multimodal reasoning is highly sensitive to these specific heads, highlighting their non-redundant necessity. 
Crucially, this localized functional sparsity paves the way for efficient model optimization, suggesting that operating on a small fraction of critical heads can significantly accelerate inference while maintaining precise semantic selection.

%% file: paper/related_work.tex
\section{Related Work}
\label{sec:related_work}

\subsection{Mechanistic Interpretability of MLLMs}
\label{subsec:Mechanistic Interpretability of VLMs}
Recent mechanistic interpretability research on MLLMs shows that their perceptual and reasoning capabilities rely on specialized and sparsely activated attention heads.
Statistical and structural analyses of cross-modal attention suggest that certain heads exhibit functional preferences.
Recent work suggests that certain attention heads in MLLMs are associated with functional roles such as visual perception and spatial reasoning~\citep{ma2026attention}.
Quantitative analyses based on response scoring~\citep{wang2025sparsemm}, signal-based methods~\citep{basile2025head}, and entropy measures~\citep{ma2026dear} suggest that a subset of heads contributes disproportionately to task-specific adaptation and visual representation encoding.
Furthermore, causal intervention techniques, including activation patching and representation editing, have been used to investigate candidate functional circuits (e.g., visual counting~\citep{che2026counting}) and hierarchical interactions between attention heads and feed-forward networks~\citep{li2026causal}.
These mechanistic insights also enable training-free interventions to improve model reliability and efficiency.
For instance, prior studies have investigated interventions such as causal head modulation~\citep{wang2025v}, reweighting attention from visual sinks to informative heads~\citep{kang2025see}, and exploiting abnormal attention patterns for zero-shot hallucination detection~\citep{zhang2025dhcp}. 
These methods demonstrate potential benefits in improving VQA performance, mitigating hallucinations, and reducing inference cost.
Despite these advances, existing analyses are largely limited to static datasets. limiting their ability to capture fine-grained attention patterns in dynamic or densely structured visual environments.

\subsection{Cross-Modal Information Retrieval}
\label{subsec:Cross-Modal Information Retrieval}

Current MLLMs exhibit strong performance in complex multimodal reasoning tasks such as multi-hop question answering~\citep{Lim2025VLMTVM}. 
However, the mechanisms underlying their ability to locate and extract task-relevant visual features in dense visual contexts remain poorly understood. 
Existing literature primarily investigates this cross-modal information retrieval through two distinct analytical lenses: head-level structural sparsity and token-level interpretability. 
For the former, inspired by text-only LLMs~\citep{wu2025retrieval}, researchers have identified dedicated subsets of attention heads that naturally govern visual grounding and localization behaviors~\citep{kang2025your, xie2026head}. 
Parallel to these macro-level insights, finer-grained token-level approaches attempt to trace autoregressive generation directly to specific visual regions~\citep{chen2026mllmsattendrelyon, liang2025explaining} and employ explicit mechanisms like Vision-Guided Attention to anchor generated text to tangible visual cues~\citep{zhao2025tell}. 
Specifically, these token-level strategies typically compute semantic alignment scores between textual instructions and visual patches, forcing the model to assign higher importance to relevant objects. 
While effective for standard visual contexts, these techniques predominantly rely on static structural priors or post-hoc attributions. 
Consequently, existing approaches struggle to perform robust cross-modal information retrieval in modern, dense, and dynamic datasets, where target features are deeply embedded and heavily obscured by complex, evolving background distractors.

%% file: paper/3-method.tex
\section{Isolating CoRe Heads via Retrieval Attention Mass}
\label{sec:method}


To systematically isolate the attention heads responsible for precise visual retrieval and filtering background noise, we extend the text-only probing framework of QRhead~\citep{zhang25qrhead} to multimodal settings. 
Our key idea is simple: a useful head should place more attention from the query tokens onto the \emph{relevant visual regions}, not just anywhere in the sequence. Based on this intuition, we define a token-level metric called \textbf{Retrieval Attention Mass (RAM)} to quantify how much attention a head allocates to the target visual content.
For each attention head $h$, we define its RAM score as:
\begin{equation}
\label{eq:ram_score}
\mathcal{M}_{\text{RAM}}^{(h)}(q \rightarrow V^*) = \mathbb{E}_{x \in q} \left[ \sum_{y \in \Omega(V^*)} \mathbf{A}^{(h)}_{x \to y} \right]
\end{equation}

where $q$ represents the set of instructional query tokens, and $\Omega(V^*)$ denotes the tokenized spatial or temporal span corresponding to the key visual entities $V^*$. 
The term $\mathbf{A}^{(h)}_{x \to y}$ is the attention weight from token $x$ to token $y$ in head $h$.
Intuitively, this metric measures how strongly a head retrieve information from the query to the relevant visual regions. By averaging over all query tokens, RAM reflects the expected retrieval strength of a head. 
Importantly, since we restrict the summation to the ground-truth region $\Omega(V^*)$, high RAM values indicate targeted retrieval, rather than generic attention accumulation. This allows us to distinguish true retrieval heads from attention sinks that absorb probability mass without contributing a meaningful cross-modal grounding.


\begin{figure}[htbp]
    \centering
    \includegraphics[width=\linewidth]{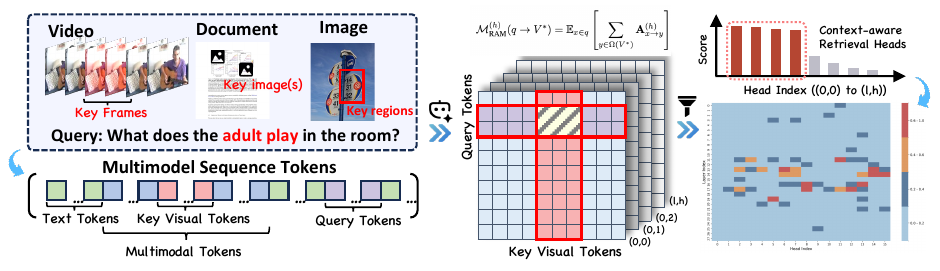}
    \caption{Overview of the CoRe head probing pipeline. The input multimodal sequence is partitioned into general text tokens, key target visual tokens ($t_v$), and instructional query tokens ($t_q$). During MLLM inference, we extract the internal self-attention maps. By aggregating the attention weights directed from $t_q$ to $t_v$, we compute a routing metric ($\mathcal{M}_{\text{RAM}}^{(h)}$) to isolate the sparse CoRe heads responsible for cross-modal information retrieval.}
    \label{fig:main}
\end{figure}


As shown in Figure~\ref{fig:main}, we first partition the multimodal input into three parts: text context, query tokens, and key visual tokens. During a standard forward pass, we extract the full self-attention maps $\mathrm{Softmax}(\frac{QK^\top}{\sqrt{d}})$ across all layers and heads.
We then focus only on the \emph{text-to-vision} attention, i.e., the sub-matrix where query tokens attend to visual tokens. This step removes irrelevant intra-modal interactions (e.g., text-to-text or vision-to-vision attention), significantly reducing noise.
By applying Equation~\eqref{eq:ram_score}, we aggregate these attention values into a scalar RAM score for each head. Collectively, this produces a global $\mathcal{M}_{\text{RAM}}^{(h)}$ distribution, where only a small subset of heads exhibit high scores. We define these high-scoring heads as \textbf{CoRe Heads}, as they are responsible for focused cross-modal retrieval.
Detailed implementations and pseudocode are provided in Appendix~\ref{subsec:unified_detection_algorithm}.

\input{Table/dataset}

\textbf{Unified Evaluation Protocol.}
To test whether CoRe heads generalize across tasks, we design a unified evaluation protocol covering four diverse multimodal datasets (see Table~\ref{tab:dataset} and Appendix~\ref{sec:appendix_dataset} for detailed descriptions). 
To capture different types of retrieval, we define the target $V^*$ differently for each task: bounding boxes for spatial grounding (RefCOCOg), spatio-temporal tubes for video grounding (VidSTG), multiple evidence regions for multi-hop reasoning (MMLongBench), and document regions for multimodal retrieval (MMDocIR). 
Since these annotations are continuous (in space or time) but the model operates over discrete tokens, we map each target region to its corresponding token indices $\Omega(V^*)$. The details of this mapping are provided in Appendix~\ref{sec:Token Mapping Rules}. This unified formulation allows us to evaluate whether CoRe heads implement a shared retrieval mechanism across modalities and tasks.

\paragraph{Generalization Across Architectures and Scales}
Finally, to verify that CoRe heads are not tied to a specific architecture, we conduct experiments across multiple model families, including Qwen3-VL, LLaVA-OneVision, and InternVL3.5. We also perform a scaling study using different sizes of Qwen3-VL (4B, 8B, 32B). 
This combined analysis enables us to examine how the sparsity and stability of CoRe heads evolve with model design and scale, and whether they reflect an intrinsic property of multimodal transformers.

%% file: Table/dataset.tex

\begin{table*}[htbp]
    \centering
    \caption{CoRe Heads Detection Experimental Configuration}
    \label{tab:dataset}
    \renewcommand{\arraystretch}{1.25}
    \resizebox{\textwidth}{!}{
        \begin{tabular}{@{} l l l l @{}}
            \toprule[1.2pt]
            \textbf{Dataset} & \textbf{Modality} & \textbf{Task Type} & \textbf{Key Visual Entity $V^*$} \\
            \midrule[0.8pt]
            RefCOCOg~\citep{kazemzadeh-etal-2014-referitgame}    & \textit{Static Image}  & Visual Grounding            & Bounding box $[x,y,w,h]$              \\
            VidSTG~\citep{zhang2020does}      & \textit{Video}         & Spatio-temporal Grounding   & Temporal tube $\mathcal{T}_{t_1:t_2}$ \\
            MMLongBench~\citep{wang2025mmlongbenchbenchmarkinglongcontextvisionlanguage} & \textit{Long Document} & Multi-evidence Reasoning    & Evidence regions $\{R_k\}$            \\
            MMDocIR~\citep{dong2025mmdocir}     & \textit{Doc + Image}   & Document-grounded Retrieval & Relevant page/figure region           \\
            \bottomrule[1.2pt]
        \end{tabular}
    }
\end{table*}

%% file: paper/4-exp.tex
\section{Mechanistic Analysis of CoRe heads}
\label{sec:experiment}

\begin{figure}[t]
\centering
\includegraphics[width=\linewidth]{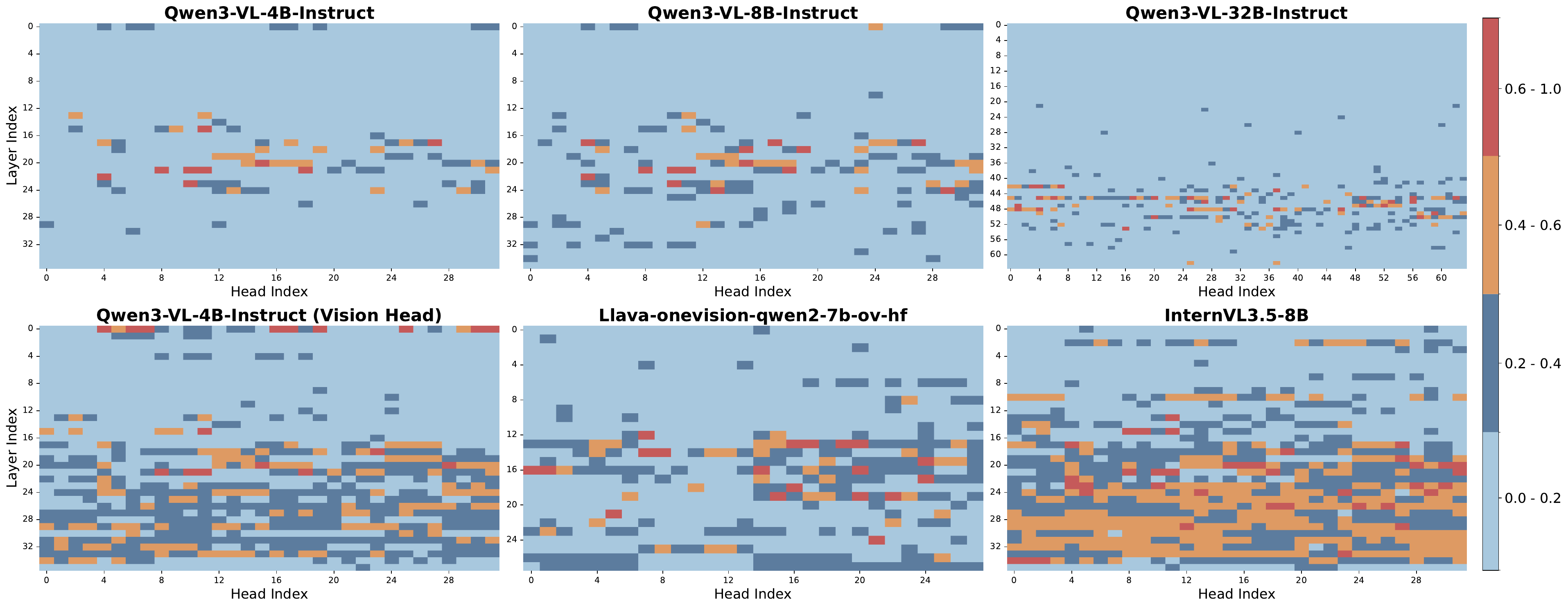}
\caption{Evolution and structural divergence of CoRe heads on the MMDocIR dataset. As model scale increases, attention patterns shift from broadly distributed activations (4B) to a pronounced deep-layer bottleneck (32B). Cross-architecture comparisons further reveal distinct attention topologies: Qwen3-VL exhibits sparse localization, Llava-onevision shows moderate dispersion, and InternVL3.5 presents dense, widespread activations. }
\label{fig:attention_map}
\end{figure}

\subsection{Distribution Dynamics across Architectures and Scales}
\label{subsec:exp_settings1}


\paragraph{Scaling Patterns within Model Families.}

As illustrated in Figure~\ref{fig:attention_map}, model capacity expansion triggers a transition from uniform dispersion to pronounced structural localization. 
In smaller variants (4B), active CoRe heads are relatively scattered across the middle layers.
However, as the scale reaches 32B, these critical heads converge into a highly concentrated, contiguous block within the deep layers (approximately layers 44–56). 
This evolution indicates that larger models spontaneously develop a more specialized "bottleneck" for cross-modal semantic integration, delegating complex visual-linguistic retrieval to a compact subset of deep-layer heads.

\paragraph{Architectural Divergence across Model Families.}
Figure~\ref{fig:attention_map} reveals stark contrasts in attention allocation across models. 
Qwen3-VL exhibits a highly sparse and localized activation structure, whereas Llava-onevision demonstrates a moderately dense and vertically dispersed pattern. 
Conversely, InternVL3.5 displays pervasive, high-density activations spanning nearly the entire network depth. 
We attribute these structural discrepancies to three primary factors:
(1) \textbf{Vision Encoder Capability:} InternVL3.5 employs a massive encoder to achieve deep semantic pre-alignment before projection, allowing the LLM to process visual tokens as "native" semantic units with high-density, layer-wise interaction. Lighter encoders shift the modality-bridging burden to the LLM, forcing the emergence of specialized deep-layer hubs.
(2) \textbf{Feature Alignment Paradigm:} The multi-stage tuning of Llava-onevision fosters progressive fusion, while the native joint optimization in Qwen3-VL induces sparse specialization to preserve linguistic reasoning capacity.
(3) \textbf{Base Model Adaptation:} Notably, despite sharing the Qwen backbone, Qwen3-VL and Llava-onevision exhibit distinct topologies. This confirms that the structural sparsity of multimodal integration is not an intrinsic property of the base LLM, but a dynamic allocation strategy responsive to the density of visual representations and the specific alignment objectives.

\subsection{Functional Decoupling: Semantic vs. Global Processing}
\label{subsec:exp_settings2}

\paragraph{Experimental Setting}

To further elucidate the functional heterogeneity among internal attention components, we evaluate and contrast two distinct categories of attention heads: 
(1) \textbf{CoRe heads}, which are systematically isolated using our proposed cross-modal attention allocation metric, 
and (2) \textbf{Vision Heads}, which are heuristically identified through the direct statistical aggregation of the macroscopic attention mass directed from linguistic query tokens to the entire set of visual tokens.

\paragraph{Result analysis}

From a distributional perspective, the two categories exhibit markedly distinct structural characteristics. 
As illustrated in the top-left panel of Figure~\ref{fig:attention_map}, the activation profiles of CoRe Heads are highly sparse and localized. 
These heads govern crucial cross-modal information selection, selectively executing the extraction of key visual features within specific layers. 
In contrast, the attention distribution of the baseline Vision Heads (depicted in the lower panel) is continuous and diffuse.
It exhibits robust responses across the middle-to-late layers with substantially broader coverage along the head dimension.
This indicates that their functionality is predominantly oriented toward global visual information aggregation, facilitating the holistic encoding and propagation of visual signals. 
This phenomenon reveals the spontaneous emergence of a structured functional decomposition mechanism within the network. 
Rather than relying on homogeneously distributed attention computations, MLLMs achieve efficient cross-modal retrieval through a critical minority of specialized heads, while leveraging a vast ensemble of auxiliary heads to ensure the stable propagation and integration of multimodal information.

\subsection{Cross-Domian Stability of CoRe heads}
\label{subsec:exp_settings3}

\begin{figure*}[htbp]
\centering
\begin{subfigure}[b]{0.48\textwidth}
\centering
\includegraphics[width=0.92\linewidth]{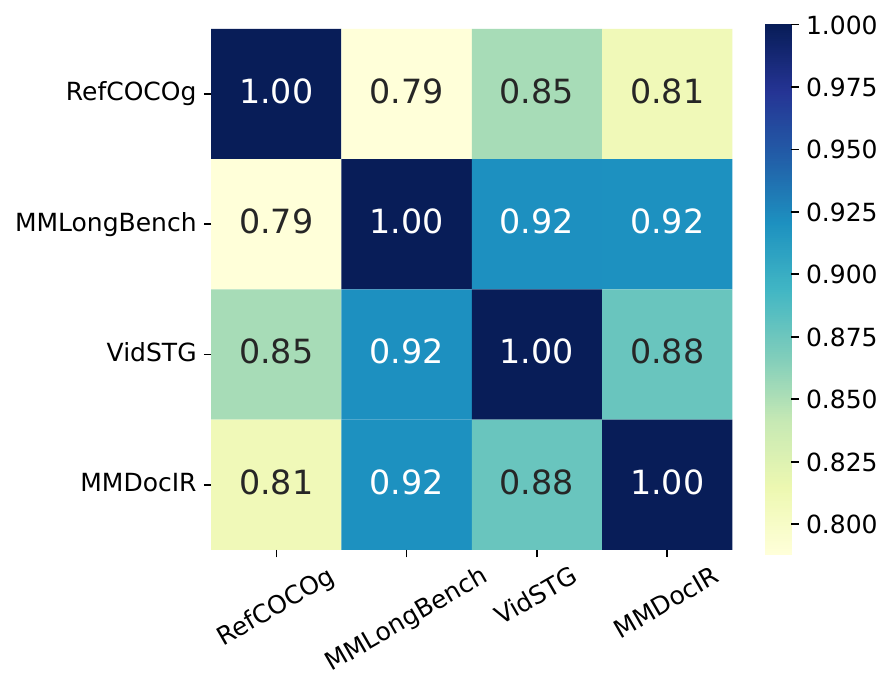}
\caption{Spearman correlation of head activations.}
\label{fig:attention_corr}
\end{subfigure}
\hfill
\begin{subfigure}[b]{0.48\textwidth}
\centering
\includegraphics[width=0.92\linewidth]{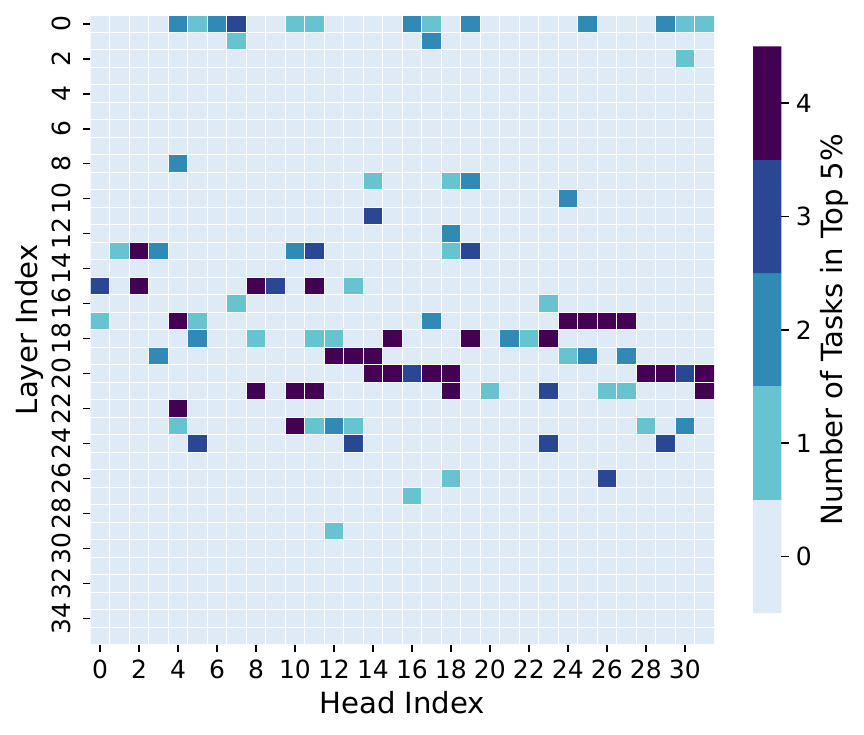}
\caption{Spatial distribution of highly activated heads.}
\label{fig:attention_stability}
\end{subfigure}
\caption{
Stability of attention heads across multi-modal tasks.
(a) The Spearman rank correlation across distinct datasets exhibits high consistency, indicating a shared information retrieval mechanism.
(b) The layer-head stability matrix illustrates the distribution of task-agnostic anchor heads. Heads consistently ranked in the top 5\% across all tasks are localized in the middle-to-late layers.
}
\label{fig:attention_analysis}
\end{figure*}

\paragraph{Experimental Setting}

To investigate the stability of CoRe heads across heterogeneous multimodal distributions, we extract head-level activation scores from Qwen3-VL-4B on RefCOCOg, MMLongBench, VidSTG, and MMDocIR. 
We measure cross-task consistency by computing pairwise Spearman rank correlations between flattened head activation vectors, capturing the agreement in relative head importance across tasks. 
And for each task, we identify heads within the top 5\% of activation scores and aggregate their occurrence frequency across all benchmarks. These frequencies are then mapped onto a layer–head grid to visualize the cross-task persistence of salient heads.Detailed implementations and pseudocode are provided in Appendix~\ref{sec:appendix_stability_details}.

\paragraph{Result analysis}

As illustrated in Figure~\ref{fig:attention_analysis}, the identified CoRe heads exhibit consistent global structural characteristics across diverse datasets. 
As depicted in Figure~\ref{fig:attention_corr}, strong positive correlations are observed across all task pairs, with coefficients ranging from 0.79 to 0.92. 
Notably, tasks demanding complex structural understanding or long-context reasoning (e.g., MMLongBench and VidSTG) demonstrate an exceptionally high correlation (0.92). 
This pronounced correlation suggests that the model universally repurposes a highly overlapping set of specialized attention heads to execute cross-modal alignment across varied downstream applications.
Furthermore, as shown in Figure~\ref{fig:attention_stability}, the majority of CoRe heads are stably concentrated within the middle layers of the architecture, predominantly distributed between layer 13 and layer 24. 
Approximately 30 specific heads remain highly activated across all four tasks, with only a marginal fraction of heads exhibiting dataset-specific activation variance. 
Consequently, the CoRe heads exhibit a "globally consistent, locally variant" distributional paradigm. 
While the core CoRe head topology demonstrates robust cross-domain generalizability, its precise activation patterns undergo adaptive modulation contingent on the target data distribution.
Detailed heatmaps for various configurations are provided in Appendix~\ref{sec:appendix_extended}.

\subsection{Causal Impact and Information Sparsity}
\label{subsec:exp_settings4}

\paragraph{Experimental Setting}

To evaluate the functional role and visual information extraction efficiency of the CoRe heads, we conduct both intervention and quantitative analyses on the Qwen3-VL-4B and Llava-onevision architectures.
For causal validation, we perform attention head masking during inference by ablating $k \in {5, 10, 20, 30}$ heads and measuring the resulting performance degradation on MMLongBench. 
We consider three ablation strategies: masking the Top-$k$ heads ranked by $\mathcal{M}_{\text{RAM}}^{(h)}$, masking the Bottom-$k$ heads, and randomly masking $k$ heads.
In parallel, to assess visual information extraction efficiency, we introduce two metrics: Key Token Ratio, which evaluates the precision of individual heads by measuring the overlap between their top 5\% attended tokens and ground-truth critical visual patches, and Key Token Coverage, which measures the collective coverage of such patches by aggregating the top attended tokens across head groups.Detailed implementations and pseudocode are provided in Appendices~\ref{sec:appendix_intervention}, \ref{sec:appendix_key_token_ratio}, and \ref{sec:appendix_key_token_coverage}.

\begin{figure*}[t]
\centering
\begin{subfigure}[b]{0.32\textwidth}
\centering
\includegraphics[width=1\linewidth]{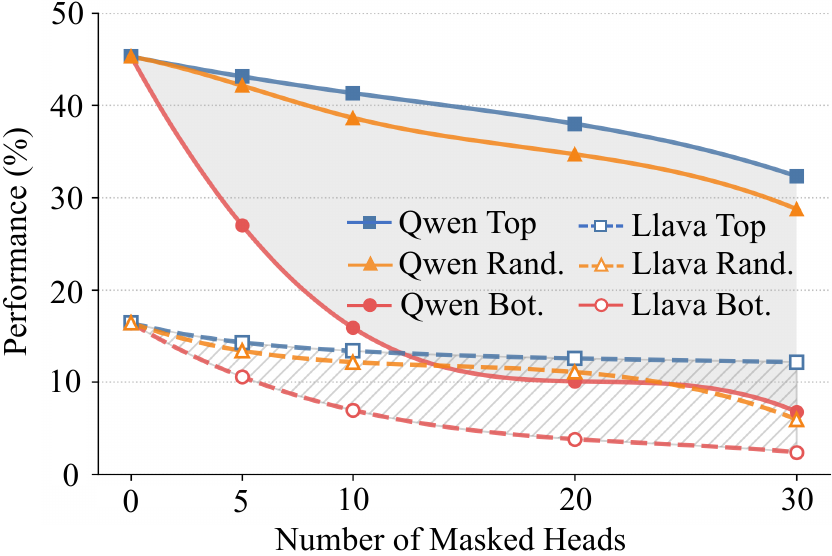}
\caption{Causal impact of head masking.}
\label{fig:mask_ablation}
\end{subfigure}
\hfill
\begin{subfigure}[b]{0.32\textwidth}
\centering
\includegraphics[width=1\linewidth]{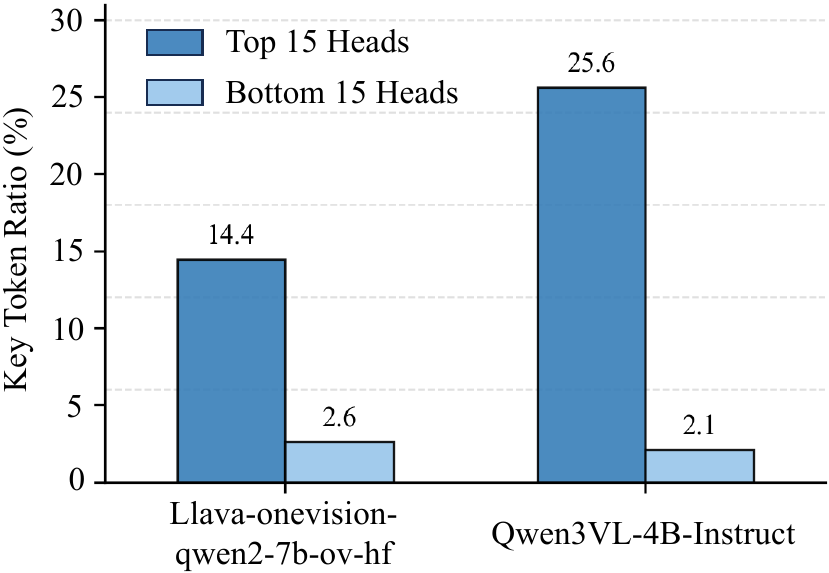}
\caption{Key visual token concentration.}
\label{fig:head_ablation_ratio}
\end{subfigure}
\hfill
\begin{subfigure}[b]{0.32\textwidth}
\centering
\includegraphics[width=1\linewidth]{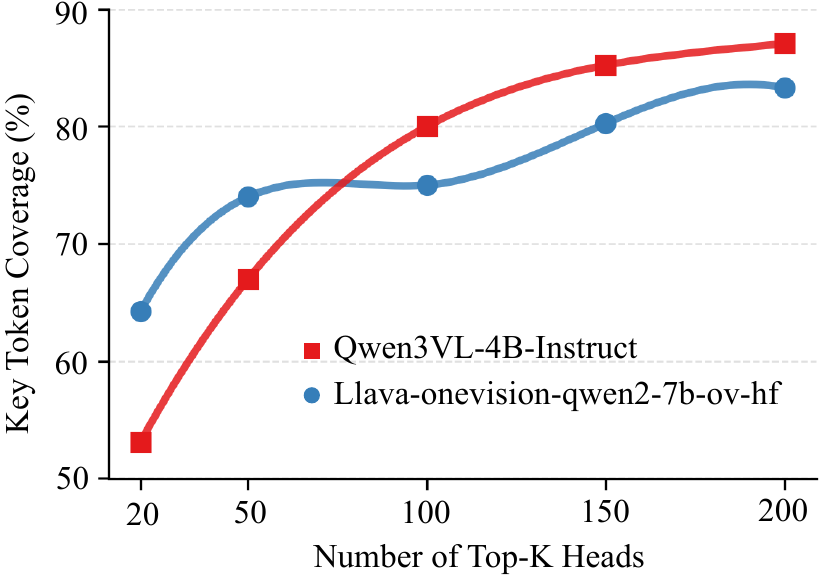}
\caption{Rapid token coverage saturation.}
\label{fig:head_ablation_coverage}
\end{subfigure}
\caption{
Quantitative analysis of the causal impact and structural sparsity of CoRe heads in MLLMs. (a) Performance degradation under head masking reveals that top-ranked CoRe heads are causally indispensable for complex multimodal reasoning tasks, whereas bottom-ranked heads yield minimal impact. (b) Key token ratios demonstrate that critical visual semantics are highly concentrated within a compact subset of elite heads. (c) The rapid saturation of cumulative token coverage confirms the structural sparsity of localized cross-modal feature extraction.
}
\label{fig:attention_ablation}
\vspace{-1em}
\end{figure*}


\paragraph{Assessing Causal Impact via Attention Head Intervention}

As illustrated in Figure~\ref{fig:mask_ablation}, a pronounced performance divergence among the different ablation strategies emerges as the number of masked heads increases. 
Masking the Top-$k$ CoRe heads (red lines) induces a precipitous and catastrophic degradation in model performance. 
For instance, ablating merely the top 5 heads causes the Accuracy of the Qwen model to plummet from approximately 45.3 to 27.0, and that of the Llava model from 16.4 to 10.6. When the mask size expands to $k=30$, the multimodal comprehension capabilities of both models effectively collapse, with accuracy scores dropping below 7. 
This phenomenon strongly validates that the identified CoRe heads govern critical information aggregation and cross-modal retrieval mechanisms during inference. 
Conversely, masking the Bottom-$k$ heads (blue lines) yields only a marginal performance drop, exhibiting an ablation trajectory that is more robust than the random masking baseline. 
This performance gap demonstrates that the top-ranked CoRe heads are not merely correlated with multimodal semantic integration; rather, they are causally indispensable for complex vision-language understanding tasks. 
They function as a highly concentrated informational bottleneck within the network architecture. 
In contrast, heads with lower importance scores contribute minimally to cross-modal feature interaction, indicating a significant degree of functional redundancy.


\paragraph{Emergent Sparsity in Multimodal Models}

As illustrated in Figure~\ref{fig:head_ablation_ratio}, the Top 15 CoRe heads concentrate a remarkably high proportion of critical visual tokens, reaching 14.4\% and 25.6\% for the Llava and Qwen models, whereas the Bottom 15 heads capture a negligible fraction of approximately 2\%. 
This contrast validates that the top-ranked CoRe heads function as highly efficient, localized hubs for cross-modal feature interaction, selectively distilling salient visual semantics. 
Figure~\ref{fig:head_ablation_coverage} further reveals that the cumulative coverage trajectories for both models exhibit a rapid initial ascent followed by swift saturation. 
Notably, aggregating merely the top 50 to 100 heads is sufficient to encompass 70\% to 80\% of the critical visual tokens. 
This rapid saturation phenomenon provides compelling quantitative evidence for the structural sparsity of multimodal retrieval. 
It indicates that the MLLMs avoids distributing critical visual processing uniformly; instead, it delegates core feature extraction mechanisms to a remarkably compact subset of elite CoRe heads.

%% file: paper/5-analysis.tex
\input{Table/MLVU}
\section{System-Level Acceleration via CoRe head-Guided Sparsity}
\label{sec:analysis}

\subsection{Methodology and Experimental Setup}
\label{subsec:Experimental Setup for Attention Allocation Strategy}

\paragraph{CoRe head-Guided Hybrid Attention Paradigm}

To empirically validate the structural sparsity of CoRe heads and their potential for accelerating inference, we implement a head-level hybrid attention strategy. 
Conventional MLLMs inherently suffer from quadratic computational complexity during the prefill phase of long visual contexts. 
Motivated by our finding that cross-modal semantic integration is highly concentrated within a sparse subset of CoRe heads, we apply a deterministic attention allocation mask to mitigate this bottleneck. 
Specifically, during the prefill stage, we inject a static head configuration into the attention forward pass. 
For the top-$k$ critical CoRe heads, we retain the standard Full Attention formulation to preserve their capacity for dense, global multimodal semantic extraction. Conversely, for the remaining non-essential vision heads, we fallback to a Stream Sparse Attention mechanism~\citep{xiao2024efficient}, restricting their computations strictly within localized sliding windows.
Detailed implementations and pseudocode are provided in Appendices~\ref{subsec:core_guided_hybrid_attention} and~\ref{subsec:acceleration_implementation}.

\paragraph{Baselines and Configurations}

We systematically ablate the proportion of attention heads retained for Full Attention computation.
Specifically, we vary the proportion of top-ranked CoRe heads that are retained in a dense attention state, while all remaining heads are constrained to operate under Stream Sparse Attention. 
Our evaluations are conducted across the Qwen3-VL (8B and 32B) and Llava-onevision architectures, establishing the standard, unmodified models (utilizing 100\% Full Attention) as our Dense baselines. 
To precisely monitor the performance-efficiency trade-offs across different cognitive granularities, we comprehensively evaluate the models on diverse subsets of the MLVU benchmark~\citep{Zhou2024MLVUBM}, which effectively decouples multimodal comprehension into holistic tasks, single-detail reasoning, and multi-detail spatiotemporal perception.
Due to space constraints, the evaluation results on the VideoMME benchmark~\cite{tang2025video} are deferred to Appendix~\ref{subsec:videomme}, which exhibit a consistent trend with those observed on MLVU.

\subsection{Results and Analysis}
\label{subsec:Results and Analysis}

\paragraph{Efficacy of System-Level Acceleration}

\begin{wrapfigure}{r}{0.45\linewidth}
  \centering
  \vspace{-1em}
        \includegraphics[width=\linewidth]{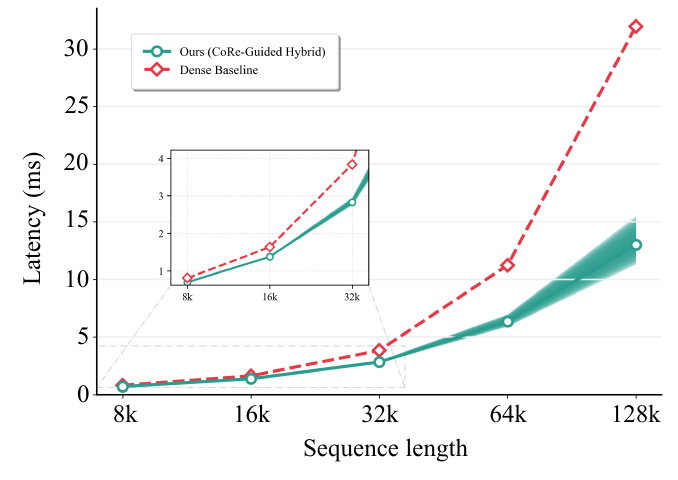}
    \caption{Our CoRe-Guided Hybrid approach consistently achieves lower latency compared to the dense baseline(Qwen3-VL-8B), with the gap widening as sequence length increases, demonstrating better scalability for long sequences. The inset highlights performance in the short-sequence regime.}
    \label{fig:qwen3vl_8b_scaling}
    \vspace{-1em}
\end{wrapfigure}

Table~\ref{tab:model_performance_updated} presents the quantitative evaluation of our CoRe head-guided hybrid attention paradigm across multiple model architectures. 
The empirical results demonstrate that dense global attention is highly redundant for the majority of attention heads. 
Across all evaluated models, our proposed structural sparsity mechanism consistently achieves significant prefill speedups. 
Remarkably, under specific configurations such as retaining $19.1\%$ of the CoRe heads in Llava-onevision and $24.4\%$ in Qwen3-VL-8B, this hybrid paradigm not only realizes a $1.8\times$ acceleration but also marginally outperforms the fully Dense baselines in overall average performance. 
These findings substantiate our hypothesis that cross-modal information retrieval is highly concentrated within a critical subset of CoRe heads; strictly pruning the receptive fields of non-essential attention heads effectively circumvents computational bottlenecks without compromising the representational integrity of the model.
As shown in Figure~\ref{fig:qwen3vl_8b_scaling}, the latency advantage becomes increasingly pronounced as sequence length grows, highlighting the scalability of our approach.

\paragraph{Granular Impact on Multimodal Comprehension}

A detailed analysis across tasks of varying cognitive granularities within the MLVU benchmark reveals task-specific model behaviors under the sparse paradigm. 
For holistic reasoning and broad detail extraction tasks, The introduction of the hybrid attention mechanism results in the performance drop typically remains within a minimal range of 1\% to 3\%, even under aggressive sparsity settings (e.g., retaining fewer than 5\% of attention heads). 
Conversely, for precise single-detail reasoning (ER) and complex multi-detail spatiotemporal perception tasks (AO, AC), the introduction of structural sparsity frequently yields performance improvements. 
We attribute this counterintuitive phenomenon to an implicit regularization effect: restricting the computation of non-essential attention heads to localized sliding windows intrinsically filters out cross-modal noise in long sequences, thereby enhancing the model's spatiotemporal focus on precise visual cues.

\paragraph{Robustness Across Model Scales}

Furthermore, as model capacity increases, CoRe head-guided attention exhibits improved robustness under sparsity constraints. 
Comparing the 8B and 32B variants of Qwen3-VL, we observe that larger models can sustain substantially higher levels of head sparsification with only minor performance degradation. 
In particular, Qwen3-VL-32B remains effective even when dense computation is reduced to 4.9\% of attention heads, achieving a 2.1× inference speedup with an average performance drop of only 0.7 points. 
These results suggest that scaling enhances redundancy in attention allocation, thereby allowing more aggressive yet stable removal of non-critical heads. 
Consequently, our deterministic head masking strategy enables controlled isolation of task-relevant CoRe heads and provides a simple and effective mechanism for accelerating inference in large multimodal models without significant loss in performance.

%% file: Table/MLVU.tex
\newcommand{\up}[1]{\textsubscript{\textcolor{blue!50}{\scriptsize $\uparrow$#1}}}
\newcommand{\down}[1]{\textsubscript{\textcolor{red!70}{\scriptsize $\downarrow$#1}}}
\newcommand{\same}[1]{\textsubscript{\textcolor{gray!70}{\scriptsize $\pm$0.0}}}

\definecolor{baselinecolor}{gray}{0.95}

\begin{table*}[htbp]
\centering
\renewcommand{\arraystretch}{1.2}
\caption{Performance comparison of different models across multiple tasks. Best results for each model are highlighted in \textbf{bold}. The subscript indicates the absolute difference compared to the corresponding Dense.The overall performances on MLVU dev set, including the holistic LVU tasks (TR: Topic Reasoning), the single-detail LVU tasks (NQA: Needle QA, ER: Ego Reasoning, PQA: Plot QA),
and multi-detail LVU tasks (AO: Action Order, AC: Action Count). }
\label{tab:model_performance_updated}
\setlength{\tabcolsep}{5pt}
\resizebox{\textwidth}{!}{
\begin{tabular}{l|r|cccccc|c|c}
\toprule
\multirow{2.5}{*}{\textbf{Model}} & \multirow{2.5}{*}{\textbf{Config}} & \multicolumn{1}{c}{\textbf{Holistic}} & \multicolumn{3}{c}{\textbf{Single Detail}} & \multicolumn{2}{c|}{\textbf{Multi Detail}} & \multirow{2.5}{*}{\textbf{Perf. Avg.}} & \multirow{2.5}{*}{\textbf{Speedup}} \\
\cmidrule(lr){3-3} \cmidrule(lr){4-6} \cmidrule(lr){7-8}
& & TR & NQA & ER & PQA & AO & AC & & \\
\midrule

\rowcolor{baselinecolor} 
\cellcolor{white} & Dense & 89.0 & 74.4 & 65.6 & \textbf{75.0} & 47.1 & 37.4 & 64.7 & 1.0$\times$ \\
 & 2.6\%  & 85.9\down{3.1} & 68.5\down{5.9} & 62.8\down{2.8} & 67.9\down{7.1} & 44.4\down{2.7} & 36.4\down{1.0} & 61.0\down{3.7} & \textcolor{green!50!black}{\textbf{2.1$\times$}} \\
 & 3.8\%  & 87.5\down{1.5} & 70.1\down{4.3} & 66.2\up{0.6} & 70.1\down{4.9} & \textbf{47.9}\up{0.8} & 36.4\down{1.0} & 63.0\down{1.7} & \textcolor{green!50!black}{\textbf{2.1$\times$}} \\
 & 6.4\%  & 87.8\down{1.2} & 70.4\down{4.0} & 65.3\down{0.3} & 69.6\down{5.4} & 44.8\down{2.3} & \textbf{37.9}\up{0.5} & 62.6\down{2.1} & \textcolor{green!50!black}{\textbf{2.0$\times$}} \\
 & 12.8\% & 86.7\down{2.3} & 70.7\down{3.7} & \textbf{68.2}\up{2.6} & 70.1\down{4.9} & 46.3\down{0.8} & 36.9\down{0.5} & 63.2\down{1.5} & \textcolor{green!50!black}{\textbf{1.9$\times$}} \\
 \multirow{-6.5}{*}{LLaVA-OneVis.}& 19.1\% & \textbf{89.4}\up{0.4} & \textbf{74.6}\up{0.2} & 66.8\up{1.2} & 73.1\down{1.9} & 47.5\up{0.4} & 37.4\same{} & \textbf{64.8}\up{0.1} & \textcolor{green!50!black}{\textbf{1.8$\times$}} \\

\midrule

\rowcolor{baselinecolor}
\cellcolor{white} & Dense & \textbf{90.5} & \textbf{74.1} & 57.1 & \textbf{74.6} & 61.4 & 37.4 & 65.8 & 1.0$\times$ \\
 & 1.7\%  & 89.0\down{1.5} & 68.7\down{5.4} & \textbf{59.4}\up{2.3} & 67.3\down{7.3} & 58.3\down{3.1} & 34.0\down{3.4} & 62.8\down{3.0} & \textcolor{green!50!black}{\textbf{2.3$\times$}} \\
 & 2.6\%  & 89.4\down{1.1} & 71.8\down{2.3} & 58.8\up{1.7} & 73.3\down{1.3} & 57.5\down{3.9} & 33.5\down{3.9} & 64.0\down{1.8} & \textcolor{green!50!black}{\textbf{2.3$\times$}} \\
 & 4.3\%  & 89.7\down{0.8} & 72.7\down{1.4} & 58.5\up{1.4} & 74.4\down{0.2} & 59.5\down{1.9} & 34.0\down{3.4} & 64.8\down{1.0} & \textcolor{green!50!black}{\textbf{2.2$\times$}} \\
 & 8.7\%  & 90.1\down{0.4} & 73.8\down{0.3} & 58.2\up{1.1} & 74.0\down{0.6} & \textbf{61.8}\up{0.4} & 35.4\down{2.0} & 65.6\down{0.2} & \textcolor{green!50!black}{\textbf{2.1$\times$}} \\
 & 13.0\% & 90.1\down{0.4} & 73.5\down{0.6} & 58.0\up{0.9} & 74.4\down{0.2} & 60.6\down{0.8} & 37.4\same{} & 65.7\down{0.1} & \textcolor{green!50!black}{\textbf{2.0$\times$}} \\
\multirow{-7.5}{*}{Qwen3-VL-8B} & 21.7\% & 90.1\down{0.4} & 73.5\down{0.6} & 58.0\up{0.9} & 73.7\down{0.9} & 61.0\down{0.4} & \textbf{39.3}\up{1.9} & \textbf{65.9}\up{0.1} & \textcolor{green!50!black}{\textbf{1.8$\times$}} \\

\midrule

\rowcolor{baselinecolor}
\cellcolor{white} & Dense & \textbf{89.7} & 76.9 & \textbf{63.6} & 75.0 & \textbf{72.2} & \textbf{38.8} & \textbf{69.4} & 1.0$\times$ \\
 & 1.2\%  & 89.4\down{0.3} & 76.6\down{0.3} & 62.8\down{0.8} & 74.8\down{0.2} & 68.3\down{3.9} & 33.5\down{5.3} & 67.6\down{1.8} & \textcolor{green!50!black}{\textbf{2.2$\times$}} \\
 & 2.4\%  & 89.4\down{0.3} & 76.6\down{0.3} & 61.6\down{2.0} & 75.0\same{} & 70.3\down{1.9} & 35.9\down{2.9} & 68.1\down{1.3} & \textcolor{green!50!black}{\textbf{2.2$\times$}} \\
 & 3.7\%  & 89.4\down{0.3} & \textbf{77.2}\up{0.3} & 60.5\down{3.1} & 74.6\down{0.4} & 71.0\down{1.2} & 35.9\down{2.9} & 68.1\down{1.3} & \textcolor{green!50!black}{\textbf{2.1$\times$}} \\
 & 4.9\%  & 89.0\down{0.7} & 76.6\down{0.3} & 63.4\down{0.2} & \textbf{75.5}\up{0.5} & \textbf{72.2}\same{} & 35.4\down{3.4} & 68.7\down{0.7} & \textcolor{green!50!black}{\textbf{2.1$\times$}} \\
 & 6.1\%  & 89.4\down{0.3} & 76.3\down{0.6} & 63.1\down{0.5} & 75.3\up{0.3} & 71.8\down{0.4} & 35.9\down{2.9} & 68.6\down{0.8} & \textcolor{green!50!black}{\textbf{2.1$\times$}} \\
\multirow{-7.5}{*}{Qwen3-VL-32B} & 7.3\%  & 89.0\down{0.7} & 76.9\same{} & 63.1\down{0.5} & 74.8\down{0.2} & 71.8\down{0.4} & 35.9\down{2.9} & 68.6\down{0.8} & \textcolor{green!50!black}{\textbf{2.0$\times$}} \\

\bottomrule
\end{tabular}
}
\end{table*}

%% file: paper/6-conclusion.tex
\section{Conclusion}
\label{sec:Conclusion}


In this work, we investigate the mechanistic basis of cross-modal information retrieval in Multimodal Large Language Models by identifying CoRe heads, a sparse subset of attention heads responsible for query-relevant visual extraction. 
Our analyses reveal a clear functional dichotomy: CoRe heads precisely localize relevant entities, while the vast majority of heads exhibit diffuse, global attention patterns. 
Through causal interventions, we establish the non-redundant necessity of these specific heads for robust multimodal reasoning.
Furthermore, our acceleration experiments validate the practical utility of this phenomenon, demonstrating that selectively preserving CoRe heads significantly expedites inference without compromising task performance. 
While preliminary, we hope these findings provide insights into the mechanistic interpretability of multimodal models and inspire further work on controllable efficiency.

%% file: paper/7-appendix.tex
\section{Code Availability}

The source code of CoRe-Head is publicly available at:
\url{https://github.com/aaxiyao/CoRe_Head}

\section{Dataset Statistics}
\label{sec:appendix_dataset}


To evaluate the cross-modal retrieval capabilities of CoRe heads across diverse scenarios, we select four datasets that cover static images, video sequences, dense documents, and multimodal input in long-context.

\paragraph{RefCOCOg} 
RefCOCOg~\citep{kazemzadeh-etal-2014-referitgame} is a large-scale benchmark for complex referring expression comprehension in static images. \textbf{(1) Providing high-precision visual anchors:} Its exhaustive bounding box annotations establish a reliable baseline to precisely compute attention distributions from complex queries to target regions. \textbf{(2) Introducing static spatial robustness validation:} The presence of similar distracting entities severely challenges target disambiguation, proving that CoRe heads maintain precise filtering in crowded static scenes. \textbf{(3) Encompassing highly complex semantic interactions:} Queries with rich attributes and spatial relationships validate that these heads execute fine-grained spatial reasoning rather than shallow noun matching.

\paragraph{MMDocIR} 
MMDocIR~\citep{dong2025mmdocir} targets multimodal document information retrieval in highly dense visual environments. \textbf{(1) Providing high-precision visual anchors:} Annotations of specific layout regions help isolate core evidence from massive text and chart noise. \textbf{(2) Introducing dense layout robustness validation:} The extreme visual clutter of long documents elevates the evaluation rigor, demonstrating the heads' robust anti-noise mechanisms. \textbf{(3) Encompassing highly complex semantic interactions:} Structured queries requiring cross-chart parsing verify that these heads possess deep semantic alignment capabilities for complex document topologies, moving beyond simple OCR.

\paragraph{MMLongBench} 
MMLongBench~\citep{wang2025mmlongbenchbenchmarkinglongcontextvisionlanguage} focuses on extreme long-context multimodal reasoning across massive visual sequences. \textbf{(1) Providing high-precision visual anchors:} Annotations of sparsely distributed evidence enable the precise computation of attention distributions while stripping away extensive contextual noise. \textbf{(2) Introducing long-range dependency robustness validation:} Severe attention dilution in massive sequences rigorously proves that CoRe heads sustain stable, high-recall retrieval mechanisms. \textbf{(3) Encompassing highly complex semantic interactions:} Queries demanding multi-hop reasoning confirm that these heads execute profound multi-evidence cross-modal routing rather than localized feature matching.

\paragraph{VidSTG} 
VidSTG~\citep{zhang2020does} is a large-scale benchmark for video spatio-temporal grounding, requiring dual localization within video streams. \textbf{(1) Providing high-precision visual anchors:} Its frame-level bounding box annotations provide a robust baseline to isolate background noise and precisely compute query-to-target attention. \textbf{(2) Introducing dynamic robustness validation:} The inherent spatio-temporal dynamics of videos increase evaluation rigor, proving that CoRe heads maintain stable alignment amidst evolving visual inputs. \textbf{(3) Encompassing highly complex semantic interactions:} Queries involving actions and relationships verify that these heads execute fine-grained cross-modal routing beyond simple entity matching.

\section{Token Mapping Rules}
\label{sec:Token Mapping Rules}

To measure the attention allocated to target visual regions by the CoRe heads, we map continuous spatial annotations to discrete 1D token indices within the language model sequence.

\subsection{Mapping Strategy for Spatial Visual Entities in RefCOCOg}
\label{subsec:refcocog_mapping}

For the RefCOCOg dataset, the ground-truth visual entity $V^*$ is provided as a continuous bounding box. 

Let the original image $\mathcal{I}$ have a resolution of $W \times H$, and the target object be localized by a bounding box defined by its top-left and bottom-right coordinates: $B = [x_{min}, y_{min}, x_{max}, y_{max}]$. The vision encoder partitions the image into a 2D patch grid of dimensions $W_{grid} \times H_{grid}$. We compute the scaling factors along the spatial axes as $s_x = \frac{W_{grid}}{W}$ and $s_y = \frac{H_{grid}}{H}$. The continuous coordinates of $B$ are then projected onto the discrete token grid. To ensure complete coverage of the target region without exceeding the image boundaries, the projected grid coordinates are defined as:
\begin{equation}
x'_{min} = \max(0, \lfloor x_{min} \cdot s_x \rfloor), \quad y'_{min} = \max(0, \lfloor y_{min} \cdot s_y \rfloor)
\end{equation}
\begin{equation}
x'_{max} = \min(W_{grid} - 1, \lfloor x_{max} \cdot s_x \rfloor), \quad y'_{max} = \min(H_{grid} - 1, \lfloor y_{max} \cdot s_y \rfloor)
\end{equation}

These 2D visual patches are then flattened into a 1D sequence in row-major order. For any target patch located at grid coordinates $(x', y')$ where $x' \in [x'_{min}, x'_{max}]$ and $y' \in [y'_{min}, y'_{max}]$, its relative 1D index within the visual sequence is:
\begin{equation}
i_{rel} = y' \cdot W_{grid} + x'
\end{equation}

Multimodal LLMs like Qwen-VL use a vision-language adapter to compress the visual sequence, often merging multiple adjacent patches into a single visual token. Let $c$ denote this downsampling factor (e.g., $c = 4$ for a $2 \times 2$ pooling mechanism). The effective relative index of the visual token becomes $\lfloor \frac{i_{rel}}{c} \rfloor$. Assuming the visual tokens are injected into the full \texttt{input\_ids} sequence starting at an absolute positional offset $\mathcal{O}_{vis}$, the final set of target visual token indices $V^*$ corresponding to the bounding box $B$ is:
\begin{equation}
V^* = \left\{ \mathcal{O}_{vis} + \left\lfloor \frac{y' \cdot W_{grid} + x'}{c} \right\rfloor \;\middle|\; x' \in [x'_{min}, x'_{max}], y' \in [y'_{min}, y'_{max}] \right\}
\end{equation}

This mapping ensures that the Retrieval Attention Mass (RAM) metric aggregates attention weights only over tokens representing the target entity, minimizing background noise and reflecting the spatial retrieval capability of the CoRe heads.

\subsection{Mapping Strategies for Structural Visual Entities in MMDocIR}
\label{subsec:mmdocir_mapping}

Unlike static single-image tasks, the MMDocIR dataset presents an extreme long-context challenge characterized by highly dense, interleaved multimodal sequences (e.g., dozens of document pages containing interspersed textual paragraphs, figures, and tables). Consequently, traditional spatial coordinate projection is inadequate. To precisely isolate the token indices of the ground-truth visual entity $V^*$ (e.g., a specific target figure) within massive sequences (often exceeding 100K tokens), our mapping protocols must be explicitly tailored to the MLLM's underlying visual encoding architecture. We introduce three distinct strategies: the \textit{Boundary-Tagging Protocol} for adapter-based models, \textit{Uniform Sequence Slicing} for fixed-expansion models, and \textit{Dynamic Cumulative Slicing} for variable-resolution models.

\paragraph{Boundary-Tagging Protocol (e.g., the Qwen-VL Paradigm)}
For models that dynamically compress visual sequences using Vision-Language Adapters, we structurally intervene in the context construction. Let the entire multimodal document context be represented as an interleaved sequence of elements $C = \{e_1, e_2, \dots, e_n\}$, where each element $e_i$ can be either a text block or a visual block. Given a complex query $Q$, assume the oracle evidence is located at a specific visual element $e^* \in C$.

\textbf{1. Context Tagging and Reconstruction:} During the preprocessing phase, rather than altering the native tokenization alignment, we introduce two auxiliary boundary markers, denoted as $\mathcal{T}_{start}$ and $\mathcal{T}_{end}$ (corresponding to \texttt{START\_IDS} and \texttt{END\_IDS}). The original context $C$ is reconstructed into a tagged sequence $C_{tag}$, where the target element $e^*$ is explicitly enveloped:
\begin{equation}
C_{tag} = \{e_1, \dots, \mathcal{T}_{start}, e^*, \mathcal{T}_{end}, \dots, e_n\}
\end{equation}

\textbf{2. Tokenization and Target Extraction:} The tagged sequence $C_{tag}$ is processed by the tokenizer to generate a discrete 1D sequence $S = [s_1, s_2, \dots, s_L]$. By linearly scanning $S$, we identify the absolute sequence indices of the boundary markers:
\begin{equation}
idx_{start} = \arg\max_{j} (s_j = \mathcal{T}_{start}), \quad idx_{end} = \arg\max_{j} (s_j = \mathcal{T}_{end})
\end{equation}
The final set of target visual token indices $V^*$ corresponding to $e^*$ is rigorously extracted as the enclosed sequence:
\begin{equation}
V^* = \{ j \mid idx_{start} < j < idx_{end} \}
\end{equation}

\paragraph{Uniform Sequence Slicing (e.g., the LLaVA Paradigm)}
Conversely, models like LLaVA circumvent complex spatial downsampling by deterministically expanding visual inputs into fixed-length token sequences marked by specific placeholders (e.g., the \texttt{<image>} token). For these architectures, tagging is unnecessary; instead, we rely on exact deterministic slicing.

\textbf{1. Global Visual Token Identification:} Let the ordered set of all visual inputs (e.g., document pages) in a sample be $\mathcal{I} = \{I_0, I_1, \dots, I_{N-1}\}$. We execute a global scan across the complete \texttt{input\_ids} sequence to locate the absolute sequence indices of all visual tokens. Let this ordered array be $\mathcal{P}_{all} = [p_0, p_1, \dots, p_{M-1}]$, where $M$ is the total number of visual tokens.

\textbf{2. Target Sequence Slicing:} Since the vision processor encodes each image into a fixed-length continuous chunk, the number of visual tokens allocated per image is derived as $P = \frac{M}{N}$. If the ground-truth evidence is located in a specific subset of images whose index set is $\mathcal{K}_{gt} \subset \{0, 1, \dots, N-1\}$, the target token indices $V^*$ are extracted through exact array slicing:
\begin{equation}
V^* = \bigcup_{k \in \mathcal{K}_{gt}} \left\{ \mathcal{P}_{all}[j] \;\middle|\; j \in [k \cdot P, (k+1) \cdot P - 1] \right\}
\end{equation}

\paragraph{Dynamic Cumulative Slicing (e.g., the InternVL Paradigm)}
For advanced architectures like InternVL that employ dynamic resolution preprocessing, images are adaptively partitioned into a variable number of patches based on their native aspect ratios. Consequently, the uniform slicing assumption ($P = M/N$) is mathematically invalid. 

\textbf{1. Dynamic Patch Allocation Tracking:} During the dynamic preprocessing phase, each image $I_i \in \mathcal{I}$ is partitioned into $c_i$ visual blocks. Let $T_{block}$ denote the fixed token length per block (e.g., 256 tokens). The specific number of visual tokens allocated for image $I_i$ is $P_i = c_i \times T_{block}$. We maintain an ordered array of these dynamic token lengths: $\mathcal{V}_{counts} = [P_0, P_1, \dots, P_{N-1}]$.

\textbf{2. Cumulative Offset Alignment:} To precisely isolate the tokens for a target ground-truth image $k \in \mathcal{K}_{gt}$, we must compute the cumulative offset of all preceding visual tokens within the global visual token array $\mathcal{P}_{all}$. The start offset index is defined as $O_k = \sum_{i=0}^{k-1} P_i$ (where $O_0 = 0$). The exact subset of target token indices $V^*$ is then extracted using this dynamic offset:
\begin{equation}
V^* = \bigcup_{k \in \mathcal{K}_{gt}} \left\{ \mathcal{P}_{all}[j] \;\middle|\; j \in \left[ O_k, O_k + P_k - 1 \right] \right\}
\end{equation}

\paragraph{Unified Attention Masking and Verification.}
Once the architecture-specific subset $V^*$ is accurately obtained using tagging, uniform slicing, or dynamic cumulative slicing, we utilize a dynamic key-value cache mechanism during the model's forward pass to capture the attention probability matrix. By slicing the global attention matrix exclusively at the indices belonging to $V^*$, we ensure that the Retrieval Attention Mass (RAM) metric precisely reflects the attention allocated to the target figure. These rigorous strategies intrinsically prevent index shifting, guaranteeing that our quantitative probing remains completely immune to the extreme visual clutter and structural noise pervasive in long-document contexts.

\subsection{Mapping Strategies for Page-Level Evidence in MMLongBench}
\label{subsec:mmlongbench_mapping}

Unlike fine-grained spatial grounding or interleaved layout retrieval, MMLongBench evaluates the "needle-in-a-haystack" retrieval capabilities of MLLMs across extensive document collections. The input inherently consists of a massive sequence of high-resolution, full-page images (e.g., multi-page PDF documents) seamlessly concatenated with complex user queries. The primary challenge here shifts from spatial localization to macro-level temporal/sequential isolation, as the context length frequently approaches the extreme limit of 100K+ tokens. To accurately compute the Retrieval Attention Mass (RAM) directed at specific ground-truth pages, we implement architecture-specific mapping protocols tailored to the underlying visual tokenization mechanisms: the \textit{Page-Level Boundary-Tagging Protocol} for adapter-based models and \textit{Uniform Page-Level Slicing} for token-expansion models.

Let the long document be represented as a chronologically ordered sequence of full-page visual elements $\mathcal{I} = \{I_0, I_1, \dots, I_{K-1}\}$, where $K$ denotes the total page capacity (e.g., $K=20$). Based on the natural language query $Q$, the oracle visual evidence is localized within a specific subset of target pages $\mathcal{E}^* \subset \{0, 1, \dots, K-1\}$.

\paragraph{Page-Level Boundary-Tagging Protocol (e.g., the Qwen-VL Paradigm).}
For models that employ complex vision-language pooling, we structurally intervene during the multi-turn message construction to prevent catastrophic token shifting over extreme sequence lengths.

\textbf{1. Dynamic Prompt Tagging:} We inject two specialized text tokens, $\mathcal{T}_{start}$ (\texttt{<GT\_START>}) and $\mathcal{T}_{end}$ (\texttt{<GT\_END>}), acting as explicit deterministic boundaries. The visual sequence is reconstructed such that any target evidence page $I_e$ ($e \in \mathcal{E}^*$) is tightly enveloped:
\begin{equation}
\mathcal{M}_{input} = \Big[\dots, I_{e-1}, \;\mathcal{T}_{start}, \;I_e, \;\mathcal{T}_{end}, \;I_{e+1}, \dots \Big] \oplus Q
\end{equation}

\textbf{2. Global Sequence Scanning:} The continuous page images are flattened into a massive 1D discrete token array $S = [s_1, s_2, \dots, s_L]$. We execute a linear scan across $S$ to capture the absolute index boundaries for each evidence page:
\begin{equation}
idx_{start}^{(e)} = \arg\max_{j} (s_j = \mathcal{T}_{start}), \quad idx_{end}^{(e)} = \arg\max_{j} (s_j = \mathcal{T}_{end})
\end{equation}
The ultimate set of target visual token indices is strictly defined as the union of all enclosed sequence blocks:
\begin{equation}
V^* = \bigcup_{e \in \mathcal{E}^*} \{ j \mid idx_{start}^{(e)} < j < idx_{end}^{(e)} \}
\end{equation}

\paragraph{Uniform Page-Level Slicing (e.g., the LLaVA Paradigm).}
For token-expansion models like LLaVA, structural tagging is unnecessary. Instead, these models allocate a fixed, deterministic number of tokens for each input image placeholder, allowing for precise mathematical slicing based on the model's native \texttt{image\_token\_id}.

\textbf{1. Global Vision Token Extraction:} By scanning the complete sequence of input IDs, we locate the absolute sequence positions of all visual tokens, forming an ordered array $\mathcal{P}_{all} = [p_0, p_1, \dots, p_{M-1}]$, where $M$ is the total count of visual tokens in the entire document.

\textbf{2. Deterministic Index Slicing:} Given the uniform expansion property, the fixed number of tokens allocated per document page is derived as $T_{page} = \frac{M}{K}$. To isolate the exact token indices representing the target multi-page evidence, we project the page indices $e \in \mathcal{E}^*$ onto the global vision token array:
\begin{equation}
V^* = \bigcup_{e \in \mathcal{E}^*} \left\{ \mathcal{P}_{all}[j] \;\middle|\; j \in \left[ e \cdot T_{page}, (e+1) \cdot T_{page} - 1 \right] \right\}
\end{equation}

\paragraph{Unified Target Verification.}
By extracting the architecture-specific subset $V^*$ and dynamically masking the global attention probability matrix during the model's forward pass, we isolate the query-to-evidence attention flows. These deterministic page-level tracking paradigms guarantee that our quantitative probing is completely immune to the macro-level structural noise generated by dozens of irrelevant document pages, rigorously validating the CoRe heads' capacity for long-range cross-modal routing.

\subsection{Mapping Strategies for Temporal Visual Entities in VidSTG}
\label{subsec:vidstg_mapping}

Unlike static spatial grounding or document layout analysis, the VidSTG benchmark evaluates the spatio-temporal retrieval capabilities of MLLMs in dynamic, unconstrained video streams. The input comprises a lengthy chronological sequence of sampled video frames, seamlessly concatenated with natural language queries. As the temporal context expands (e.g., up to 128 frames per video), the model encounters massive temporal background noise. To accurately compute the Retrieval Attention Mass (RAM) directed exclusively at the ground-truth action or object frames, we adapt our architecture-specific mapping protocols to the temporal domain, transitioning from page-level extraction to fine-grained temporal sequence isolation.

Let the input video be represented as a chronologically ordered set of sampled frames $\mathcal{V} = \{F_0, F_1, \dots, F_{N-1}\}$, where $N$ denotes the total number of extracted frames. Based on the temporal annotations, the oracle visual evidence corresponds to a specific continuous or discrete subset of target frames, whose index set is defined as $\mathcal{K}_{gt} \subset \{0, 1, \dots, N-1\}$.

\paragraph{Temporal Boundary-Tagging Protocol (e.g., the Qwen-VL Paradigm).}
For models employing Vision-Language Adapters that dynamically compress temporal visual tokens, we intervene at the text-prompt construction phase to prevent temporal index shifting. 

\textbf{1. Dynamic Frame Tagging:} We inject explicit temporal boundaries, $\mathcal{T}_{start}$ and $\mathcal{T}_{end}$ (corresponding to \texttt{START\_IDS} and \texttt{END\_IDS}), surrounding the specific target frames. The temporal context is reconstructed such that any target evidence frame $F_k$ ($k \in \mathcal{K}_{gt}$) is tightly enveloped:
\begin{equation}
\mathcal{M}_{input} = \Big[\dots, F_{k-1}, \;\mathcal{T}_{start}, \;F_k, \;\mathcal{T}_{end}, \;F_{k+1}, \dots \Big] \oplus Q
\end{equation}

\textbf{2. Global Sequence Scanning:} The entire video-text context is flattened into a 1D discrete token array $S = [s_1, s_2, \dots, s_L]$. A linear scan captures the absolute index boundaries for each tagged target frame:
\begin{equation}
idx_{start}^{(k)} = \arg\max_{j} (s_j = \mathcal{T}_{start}), \quad idx_{end}^{(k)} = \arg\max_{j} (s_j = \mathcal{T}_{end})
\end{equation}
The target temporal token indices $V^*$ are defined as the union of these enclosed sequence blocks:
\begin{equation}
V^* = \bigcup_{k \in \mathcal{K}_{gt}} \{ j \mid idx_{start}^{(k)} < j < idx_{end}^{(k)} \}
\end{equation}

\paragraph{Uniform Temporal Slicing (e.g., the LLaVA Paradigm).}
For token-expansion models like LLaVA-OneVision, which allocate a fixed, deterministic number of tokens for each video frame placeholder, temporal structural tagging is omitted in favor of exact mathematical array slicing.

\textbf{1. Global Frame Token Extraction:} By scanning the complete \texttt{input\_ids} sequence, we locate the absolute positions of all visual tokens, forming an ordered array $\mathcal{P}_{all} = [p_0, p_1, \dots, p_{M-1}]$, where $M$ is the total token count representing the entire video.

\textbf{2. Deterministic Temporal Slicing:} Given the uniform expansion property, the fixed token length allocated per frame is exactly $T_{frame} = \frac{M}{N}$. To strictly isolate the tokens representing the target temporal tubes, we project the target frame indices $k \in \mathcal{K}_{gt}$ onto the global array:
\begin{equation}
V^* = \bigcup_{k \in \mathcal{K}_{gt}} \left\{ \mathcal{P}_{all}[j] \;\middle|\; j \in \left[ k \cdot T_{frame}, (k+1) \cdot T_{frame} - 1 \right] \right\}
\end{equation}

\paragraph{Dynamic Cumulative Temporal Slicing (e.g., the InternVL Paradigm).}
For advanced architectures like InternVL that enforce dynamic resolution preprocessing, individual frames may be partitioned into a variable number of patches depending on their motion blur or native aspect ratios. Therefore, the uniform assumption ($M/N$) is violated, necessitating a dynamic cumulative tracking mechanism.

\textbf{1. Dynamic Patch Allocation Tracking:} During video preprocessing, each frame $F_i$ is adaptively partitioned into $c_i$ visual blocks. Assuming a fixed token length per block $T_{block}$, the total tokens allocated for frame $F_i$ is $P_i = c_i \times T_{block}$. We maintain an ordered tracking array of these dynamic lengths across the temporal axis: $\mathcal{V}_{counts} = [P_0, P_1, \dots, P_{N-1}]$.

\textbf{2. Cumulative Offset Alignment:} To precisely isolate the tokens for a target temporal frame $k \in \mathcal{K}_{gt}$, we calculate the cumulative temporal offset of all preceding frames. The start offset index is formulated as $O_k = \sum_{i=0}^{k-1} P_i$ (where $O_0 = 0$). The exact target token indices $V^*$ are extracted via this dynamically computed sliding window:
\begin{equation}
V^* = \bigcup_{k \in \mathcal{K}_{gt}} \left\{ \mathcal{P}_{all}[j] \;\middle|\; j \in \left[ O_k, O_k + P_k - 1 \right] \right\}
\end{equation}

\paragraph{Unified Temporal Verification.}
By utilizing these rigorous, architecture-aware mapping protocols to extract the precise subset $V^*$, we dynamically slice the global attention probability matrix during inference. This ensures that our RAM formulation strictly evaluates the attention routed solely to the query-relevant spatio-temporal segments. These methodologies inherently neutralize the massive temporal noise introduced by dozens of irrelevant background frames, thereby strictly validating the CoRe heads' temporal reasoning robustness.

\section{CoRe Heads: Detection Mechanisms and Model Configuration}
\label{sec:CoRe Heads: Detection Mechanisms and Model Configuration}

\subsection{Model Architectures and Hyperparameters}
\label{subsec:Model Architectures and Hyperparameters}

To strictly ensure our findings regarding CoRe heads are intrinsic structural properties rather than artifacts of specific designs, we evaluate three representative MLLM families encompassing diverse vision-language integration paradigms: adapter-based compression, deterministic token expansion, and dynamic high-resolution preprocessing. The architectural distinctions are summarized in Table~\ref{tab:model_configurations}. 

\input{Table/model}

\subsection{Unified Detection Framework for CoRe Heads}
\label{subsec:unified_detection_algorithm}

Building upon the architecture-specific mapping protocols detailed in previous sections, we abstract our empirical probing methodology into a unified, computationally efficient algorithm. The core objective is to quantitatively extract the Retrieval Attention Mass (RAM) allocated to the ground-truth visual entities across any combination of MLLM architectures and heterogeneous datasets (ranging from static bounding boxes to extreme long-context video tubes).

Algorithm \ref{alg:unified_ram} presents the complete detection framework. To circumvent the prohibitive memory overhead of full-sequence generation and backpropagation, we implement a customized \textit{Dynamic Key-Value Cache} mechanism. By running a single forward pass without computing language modeling logits, we hook directly into the intermediate transformer layers. Furthermore, the algorithm is meticulously designed to handle modern architectural variants, such as Grouped Query Attention (GQA), by explicitly repeating the Key states prior to the inner product. Ultimately, the algorithm computes a normalized attention map $\mathbf{M}_{RAM} \in \mathbb{R}^{L \times H}$, identifying the precise topological location of CoRe heads.

\begin{algorithm}[ht]
\caption{Unified Extraction Framework for Retrieval Attention Mass (RAM)}
\label{alg:unified_ram}
\textbf{Input}: Multimodal context $C$, Query $Q$, Ground-truth visual entity $E^*$, Model $M$ \\
\textbf{Output}: Layer-head attention allocation map $\mathbf{M}_{RAM} \in \mathbb{R}^{L \times H}$
\begin{algorithmic}[1]
\STATE \textbf{Phase 1: Architecture-Aware Token Mapping}
\STATE $S \leftarrow \text{UnifiedTokenizer}(C \oplus Q)$
\STATE $idx_Q \leftarrow \text{FindTokenIndices}(S, Q)$ \COMMENT{Locate query tokens}
\IF{$M$ uses Adapter-based Compression (e.g., Qwen-VL)}
    \IF{Context is dense sequence}
        \STATE $V^* \leftarrow \text{BoundaryTagging}(S, E^*, \mathcal{T}_{start}, \mathcal{T}_{end})$
    \ELSE
        \STATE $V^* \leftarrow \text{SpatialProjection}(E^*, \text{downsample\_ratio}=c)$
    \ENDIF
\ELSIF{$M$ uses Deterministic Expansion (e.g., LLaVA)}
    \STATE $V^* \leftarrow \text{UniformSlicing}(S, E^*, T_{page/frame})$
\ELSIF{$M$ uses Dynamic High-Res Preprocessing (e.g., InternVL)}
    \STATE $\mathcal{V}_{counts} \leftarrow \text{TrackPatchAllocations}(C)$
    \STATE $V^* \leftarrow \text{CumulativeOffsetSlicing}(S, E^*, \mathcal{V}_{counts})$
\ENDIF

\STATE \textbf{Phase 2: Custom Cache and Forward Pass}
\STATE Initialize $\mathcal{C}_{kv} \leftarrow \text{DynamicCacheWithQuery}(idx_Q)$
\STATE $\text{ModelForward}(M, \text{input}=S, \text{past\_key\_values}=\mathcal{C}_{kv}, \text{compute\_logits}=\text{False})$

\STATE \textbf{Phase 3: RAM Computation across Layers and Heads}
\STATE Initialize $\mathbf{M}_{RAM}$ as an $L \times H$ zero matrix
\FOR{each layer $l \in \{1, 2, \dots, L\}$}
    \STATE $Q^{(l)}, K^{(l)} \leftarrow \mathcal{C}_{kv}.\text{get\_layer}(l)$
    \STATE $K^{(l)}_{GQA} \leftarrow \text{RepeatKV}(K^{(l)}, \text{num\_kv\_groups})$ \COMMENT{Align dimensions for GQA}
    
    \STATE $\mathbf{A}^{(l)} \leftarrow \text{Softmax}\left( \frac{Q^{(l)} \left(K^{(l)}_{GQA}\right)^T}{\sqrt{d_{head}}} \right)$ \COMMENT{Global attention probability}
    
    \FOR{each head $h \in \{1, 2, \dots, H\}$}
        \STATE $\mathbf{M}_{RAM}[l, h] \leftarrow \sum_{j \in V^*} \mathbf{A}^{(l, h)}_{:, j}$ \COMMENT{Aggregate mass over target tokens}
    \ENDFOR
\ENDFOR
\STATE \textbf{return} $\mathbf{M}_{RAM}$
\end{algorithmic}
\end{algorithm}

\section{Feature Analysis and Causal Verification of CoRe Heads}
\label{sec:Feature Analysis and Causal Verification of CoRe Heads}

\subsection{Implementation Details for Cross-Domain Stability Analysis}
\label{sec:appendix_stability_details}

This section provides the implementation details for the cross-domain stability and correlation analysis of CoRe heads presented in the main text. To quantitatively assess the functional consistency of attention heads across diverse multimodal distributions, we implemented a unified evaluation pipeline across four representative datasets: RefCOCOg, MMLongBench, VidSTG, and MMDocIR.

\paragraph{Data Processing and Spearman Correlation.}
For each dataset $t \in \mathcal{T}$, we first extract the head-level Retrieval Attention Mass (RAM) scores, denoted as $\mathcal{M}_{\text{RAM}}^{(l,h)}$, for all layers $l$ and heads $h$. Missing values resulting from layer-head extraction are zero-padded to ensure uniform tensor dimensions across tasks. To measure cross-task consistency, we flatten the spatial dimensions and compute the pairwise Spearman rank correlation coefficient between the score distributions of each task pair. The resulting correlation matrix is visualized to highlight the strong positive correlations across all domains.

\paragraph{Top-$k$ Stability Masking and Spatial Distribution.}
To isolate the universally critical heads, we compute a stability matrix $S \in \mathbb{R}^{L \times H}$ that aggregates the occurrence frequency of top-ranked heads across all tasks. Specifically, for each task $t$, we calculate a dynamic threshold $\tau_t$ corresponding to the 95th percentile (top 5\%) of its RAM score distribution. We then generate a binary activation mask $\mathbb{I}_{l,h}^{(t)}$ for each head:
\begin{equation}
    \mathbb{I}_{l,h}^{(t)} = 
    \begin{cases} 
        1, & \text{if } \mathcal{M}_{\text{RAM}, t}^{(l,h)} \ge \tau_t \\
        0, & \text{otherwise} 
    \end{cases}
\end{equation}

The cross-domain stability count for a given head is computed by summing these indicator variables over all evaluating tasks:
\begin{equation}
    S_{l,h} = \sum_{t \in \mathcal{T}} \mathbb{I}_{l,h}^{(t)}
\end{equation}
where $S_{l,h} \in \{0, 1, 2, 3, 4\}$. 

\paragraph{Visualization Setup.}
The stability matrix is visualized using a custom discrete linear segmented colormap. To ensure the accurate visual representation of the discrete task counts, the colorbar visualization boundaries are shifted by $-0.5$ (spanning from $-0.5$ to $3.5$ for $N=4$ task overlaps) so that the color ticks perfectly align with the integer centers. All visualizations strictly adhere to the standard typographical guidelines, utilizing Times New Roman and STIX fonts for consistency.

\subsection{Implementation Details for Attention Head Intervention}
\label{sec:appendix_intervention}

To rigorously assess the causal impact of CoRe heads (as discussed in Section~\ref{subsec:exp_settings4}), we implemented a head-level attention intervention mechanism. The complete unified intervention procedure is formally outlined in Algorithm~\ref{alg:causal_intervention}.

\begin{algorithm}[htbp]
\caption{Memory-Efficient Chunk-wise Attention Head Intervention}
\label{alg:causal_intervention}
\textbf{Input}: Query $Q \in \mathbb{R}^{B \times H \times N \times d}$, Key $K \in \mathbb{R}^{B \times H \times M \times d}$, Value $V \in \mathbb{R}^{B \times H \times M \times d}$ \\
\textbf{Input}: Current layer index $l$, Ablation set $\mathcal{B}$, Chunk size $C$ (e.g., 256), Scaling factor $\alpha$ \\
\textbf{Output}: Intervened Attention Output $O \in \mathbb{R}^{B \times H \times N \times d}$
\begin{algorithmic}[1]
\STATE Initialize empty output tensor $O$ with shape of $Q$
\STATE Extract target heads for current layer: $\mathcal{H}_l \leftarrow \{h \mid (l, h) \in \mathcal{B}\}$
\IF{$\mathcal{H}_l = \emptyset$}
    \RETURN $\text{FlashAttention}(Q, K, V)$
\ENDIF
\FOR{$i = 0$ to $N-1$ step $C$}
    \STATE $j \leftarrow \min(i + C, N)$
    \STATE $Q_{\text{chunk}} \leftarrow Q[:, :, i:j, :]$
    \STATE $S_{\text{chunk}} \leftarrow (Q_{\text{chunk}} K^\top) \cdot \alpha$ \COMMENT{Compute pre-softmax logits}
    \STATE Apply standard causal/padding mask to $S_{\text{chunk}}$ if necessary
    \FOR{$h \in \mathcal{H}_l$}
        \STATE $S_{\text{chunk}}[:, h, :, :] \leftarrow 0.0$ \COMMENT{Ablate selective retrieval via logit neutralization}
    \ENDFOR
    \STATE $A_{\text{chunk}} \leftarrow \text{Softmax}(S_{\text{chunk}}, \text{dim}=-1)$
    \STATE $A_{\text{chunk}} \leftarrow \text{Dropout}(A_{\text{chunk}})$
    \STATE $O_{\text{chunk}} \leftarrow A_{\text{chunk}} V$
    \STATE $O[:, :, i:j, :] \leftarrow O_{\text{chunk}}$
    \STATE \textbf{Free memory:} Delete $Q_{\text{chunk}}, S_{\text{chunk}}, A_{\text{chunk}}, O_{\text{chunk}}$ 
\ENDFOR
\STATE $O \leftarrow \text{Transpose and format } O$
\RETURN $O, \text{None}$ \COMMENT{Avoid returning full attention weight matrices to save VRAM}
\end{algorithmic}
\end{algorithm}

\subsection{Calculation of the Key Token Ratio}
\label{sec:appendix_key_token_ratio}

To quantitatively evaluate the extraction precision and efficiency of the identified CoRe heads, we define a granular, token-level metric termed the \textit{Key Token Ratio}. This metric assesses whether a specific attention head successfully concentrates its highest attention mass on the exact ground-truth visual tokens required to answer the query, filtering out surrounding multimodal noise.

\paragraph{Mathematical Formulation.}
Given an input sequence $S$ of length $L$, let $Q \subset S$ denote the subset of tokens corresponding to the linguistic question, and let $V^* \subset S$ denote the subset of key target visual tokens (e.g., the specific image patches or figure tokens corresponding to the ground-truth answer). 

During the forward pass, for a specific attention head $h$ in layer $l$, we extract the post-softmax attention probability matrix. We first aggregate the attention distribution directed from the query tokens to the entire sequence by averaging over the query dimension:
\begin{equation}
\bar{A}^{(h)} = \frac{1}{|Q|} \sum_{q \in Q} A_{q \rightarrow S}^{(h)} 
\end{equation}
where $A_{q \rightarrow S}^{(h)} \in \mathbb{R}^L$ represents the attention probability distribution from a single query token $q$ to all tokens in the sequence. 

To evaluate the precision of head $h$, we define a stringent threshold by isolating only the top 5\% of tokens that receive the highest attention scores. Let $k = \max(1, \lfloor L \times 0.05 \rfloor)$. We define the highly-attended token set $\mathcal{T}_{\text{top}}^{(h)}$ as the indices of the top $k$ values in the aggregated distribution $\bar{A}^{(h)}$.

The Key Token Ratio for head $h$ is then defined as the percentage of ground-truth target tokens successfully captured within this top 5\% attended set:
\begin{equation}
\text{Key Token Ratio}^{(h)} = \frac{|\mathcal{T}_{\text{top}}^{(h)} \cap V^*|}{|V^*|} \times 100 
\end{equation}

During evaluation, if a head's Key Token Ratio meets or exceeds a predefined threshold (e.g., 50\%), it is recorded as a successful "hit". We compute the average hit rate across the selected Top-$K$ and Bottom-$K$ head populations to demonstrate the functional divergence between CoRe heads and standard vision heads.

\subsection{Calculation of the Key Token Coverage}
\label{sec:appendix_key_token_coverage}

While the Key Token Ratio (detailed in Section~\ref{sec:appendix_key_token_ratio}) evaluates the retrieval precision of individual attention heads, it does not account for the collaborative dynamics among them. To assess the collective retrieval capacity of a population of heads, we introduce the \textit{Key Token Coverage} metric. This metric quantifies the proportion of the ground-truth visual tokens that are successfully captured by the aggregate attention of a specific subset of heads (e.g., the Top-$K$ CoRe heads versus the Bottom-$K$ heads).

\paragraph{Mathematical Formulation.}
Following the previous notations, let $S$ denote the input sequence, $Q$ denote the query tokens, and $V^*$ denote the set of critical ground-truth visual tokens. For a given attention head $h$, we compute the query-aggregated 1D attention distribution $\bar{A}^{(h)}$ and extract the indices of the top $p\%$ (e.g., $p=5$) most attended tokens, denoted as $\mathcal{T}_{\text{top}}^{(h)}$.

Let $\mathcal{G}$ represent a designated group of attention heads evaluated as an ensemble (for instance, $\mathcal{G}_{\text{top}}$ representing the Top-250 heads based on RAM scores). The subset of ground-truth target tokens successfully hit by an individual head $h$ is given by the intersection $(\mathcal{T}_{\text{top}}^{(h)} \cap V^*)$. 

To determine the collective coverage of the entire head group $\mathcal{G}$, we compute the union of all successfully hit target tokens across all heads in the group. This ensures that redundant retrievals of the same visual token by multiple heads are systematically deduplicated:
\begin{equation}
\mathcal{U}_{\mathcal{G}} = \bigcup_{h \in \mathcal{G}} \left( \mathcal{T}_{\text{top}}^{(h)} \cap V^* \right) 
\end{equation}

The Key Token Coverage for the group $\mathcal{G}$ is then defined as the ratio of this union's cardinality to the total number of ground-truth visual tokens:
\begin{equation}
\text{Key Token Coverage}(\mathcal{G}) = \frac{|\mathcal{U}_{\mathcal{G}}|}{|V^*|} \times 100 
\end{equation}

\paragraph{Evaluation Protocol.}
During our evaluation on the MMDocIR dataset, we systematically compute this coverage metric for both the highest-ranked CoRe heads and the lowest-ranked baseline heads across both LLaVA-OneVision and Qwen3-VL architectures. 

By comparing $\text{Key Token Coverage}(\mathcal{G}_{\text{top}})$ against $\text{Key Token Coverage}(\mathcal{G}_{\text{bottom}})$, we quantitatively demonstrate the structural sparsity of multimodal retrieval. The results confirm that a highly compact subset of Top-$K$ heads collectively encompasses the vast majority of necessary visual evidence, whereas the cumulative receptive field of the Bottom-$K$ heads fails to align with the semantic targets. Consistent with the individual head evaluations, the target token extraction ($V^*$) and the memory-efficient sequential layer-wise caching are employed to prevent memory bottlenecks during long-context inference.

\subsection{Qualitative Visualization of Functional Dichotomy in Object Grounding}
\label{subsec:Qualitative_Visualization_RefCOCOg}

\begin{figure}[htbp]
\centering
\includegraphics[width=\linewidth]{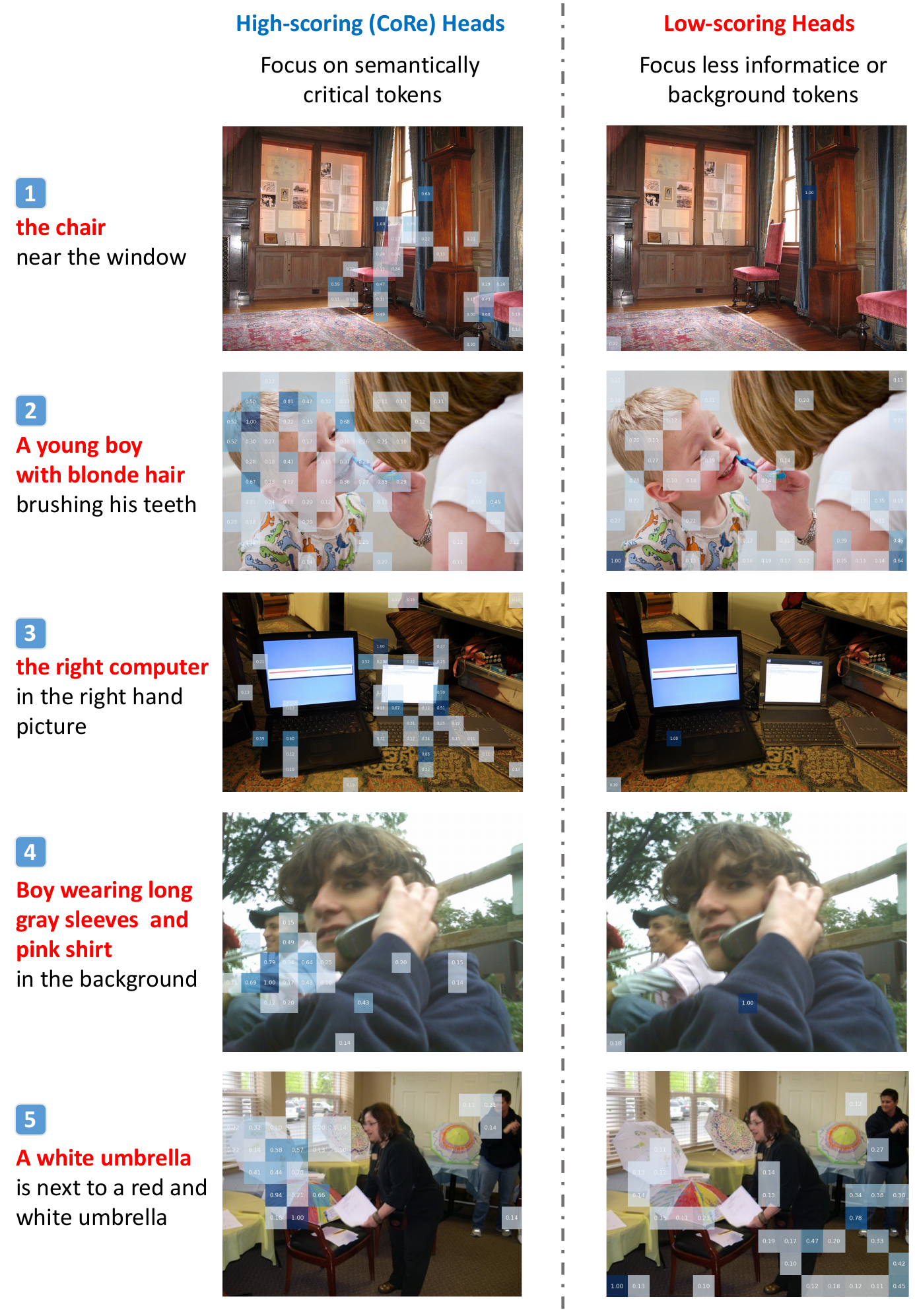}
\caption{Qualitative comparison of attention allocation on the RefCOCOg dataset. \textbf{Left (CoRe Heads):} High-scoring heads demonstrate highly precise spatial grounding, accurately isolating the visual tokens corresponding to the query-relevant entities (highlighted in red). \textbf{Right (Low-scoring Heads):} The remaining heads fail to capture task-relevant visual cues, instead distributing their attention diffusely across background noise and uninformative regions. This visually confirms the role of CoRe heads as dedicated cross-modal extractors.}
\label{fig:appendix_refcocog}
\end{figure}

Figure~\ref{fig:appendix_refcocog} provides compelling qualitative evidence of the functional dichotomy within the model's attention mechanism on the RefCOCOg dataset. As illustrated, high-scoring CoRe heads act as precise information extractors, consistently and accurately localizing the semantically critical entities specified by complex textual queries (e.g., specific clothing attributes or fine-grained spatial relationships). Conversely, the low-scoring heads exhibit severe semantic dispersion, scattering their attention mass across uninformative background regions or irrelevant objects. These stark visual contrasts directly corroborate our quantitative findings, confirming that fine-grained cross-modal grounding is exclusively governed by a highly specialized, sparse subset of attention heads rather than homogeneously distributed across the network.

\section{Extended Analysis of CoRe Attention Topologies and Model Scaling}
\label{sec:appendix_extended}

To further substantiate the observations regarding model-wide attention shifts discussed in the main text, we provide comprehensive heatmaps of the $\mathcal{M}_{\text{RAM}}^{(h)}$ (Relative Activation Magnitude) scores for the CoRe attention heads. This extended visualization contrasts the attention topologies of the Qwen3-VL series across varying parameter scales (2B vs. 4B) and task modalities, ranging from static image understanding (COCO) to complex spatio-temporal reasoning (VidSTG) and long-context document processing (MMLongBench).The empirical results consistently demonstrate that as task complexity and model scale increase, the attention distribution transitions from a broadly dispersed state to a highly localized "bottleneck structure" within the intermediate-to-deep layers. This structural divergence suggests that larger models develop specialized functional regions to manage the increased informational entropy inherent in long-range and multi-modal reasoning, effectively distilling critical cues through a narrowed architectural bottleneck.

\begin{figure}[htbp]
\centering
\includegraphics[width=\linewidth]{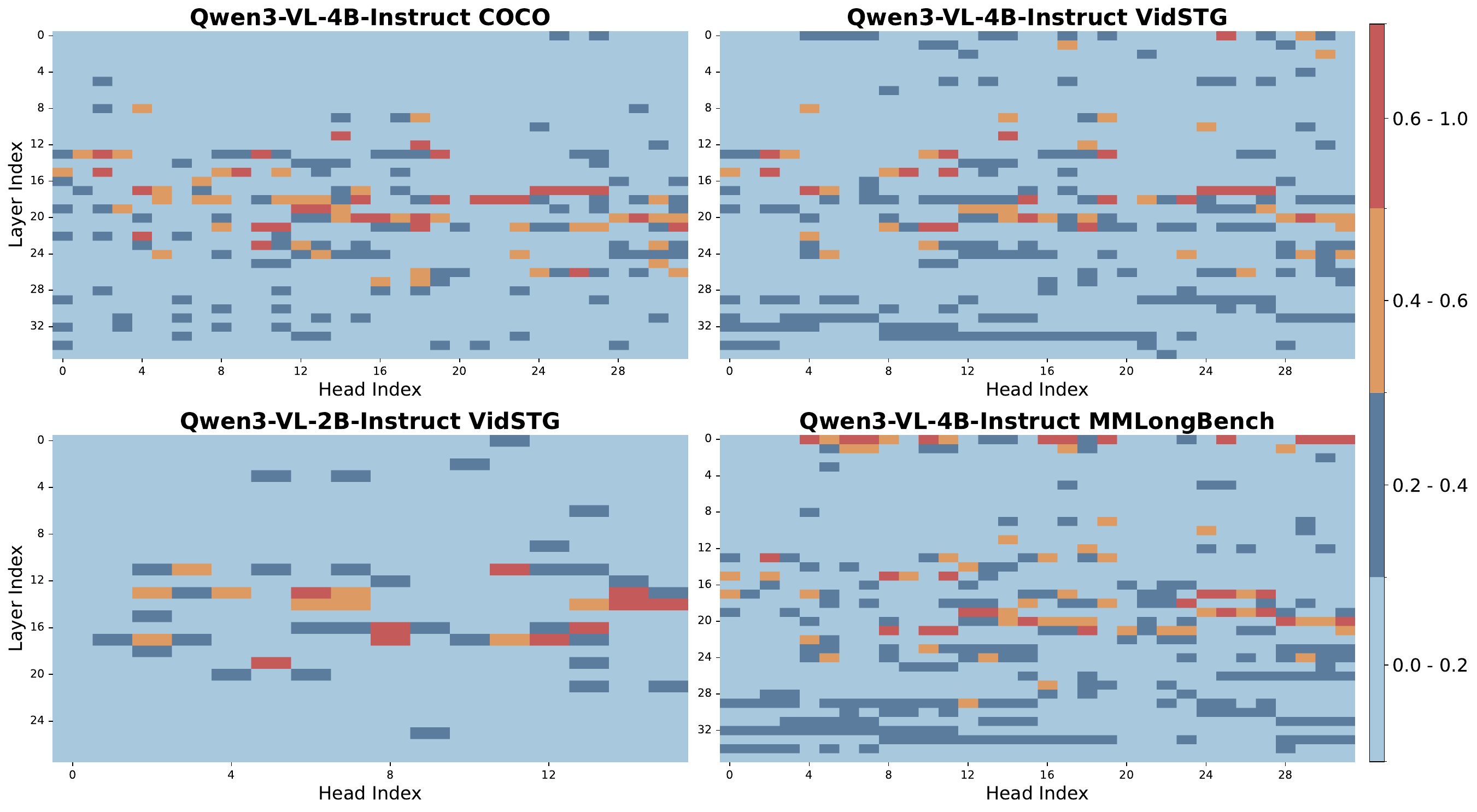}
\caption{Heatmaps of CoRe head activation distributions across heterogeneous datasets and model scales. Warmer colors indicate higher normalized activation scores. (a) Task-Specific Patterns: Contrasting the 4B model on COCO, VidSTG, and MMLongBench reveals that complex temporal and long-context tasks drive activations to cluster in specific deep layers (Layers 12–28), forming a distinct "reasoning bottleneck." (b) Scaling Effects: On the VidSTG dataset, the 4B model demonstrates a more structured, deep-layer attention concentration compared to the fragmented activations of the 2B model, indicating that increased capacity facilitates efficient feature localization.}
\label{fig:appendix_head}
\end{figure}

\section{System-Level Acceleration and Additional Benchmarks}
\label{sec:System-Level Acceleration and Additional Benchmarks}

\subsection{CoRe-Guided Hybrid Attention Mechanism}
\label{subsec:core_guided_hybrid_attention}

The quadratic computational complexity of the self-attention mechanism poses a severe bottleneck for Multimodal Large Language Models (MLLMs), particularly during the prefill stage of processing long visual contexts (e.g., high-resolution images or multi-page documents). Based on our finding that cross-modal semantic integration is highly localized within a sparse subset of CoRe heads, we introduce a system-level acceleration paradigm: \textbf{Head-level Hybrid Attention}.

As illustrated in our architectural design, we decouple the attention computation during the prefill stage based on the intrinsic importance of each attention head. The formulation is as follows:
\begin{enumerate}
    \item \textbf{Head Configuration via CoRe Ranking}: All attention heads are ranked according to their expected semantic contribution to cross-modal integration (CoRe Score). We partition the heads into two sets: the top-$k$ critical CoRe heads ($\mathcal{H}_{\text{dense}}$) and the remaining non-essential heads ($\mathcal{H}_{\text{sparse}}$).
    \item \textbf{Top-$k$ Full Attention}: For heads in $\mathcal{H}_{\text{dense}}$, we retain the standard global dense attention pattern, allowing these routing hubs to maintain unconstrained receptive fields for precise visual feature extraction.
    \item \textbf{Stream Sparse Attention}: For the vast majority of heads in $\mathcal{H}_{\text{sparse}}$, global connections are functionally redundant. We restrict their computation to a local sliding window. For a query at position $i$, these heads strictly attend to keys within a localized window $[i-w, i+w]$ alongside a small set of initial attention sinks.
\end{enumerate}
During the decoding stage (autoregressive generation), the sequence length of the query is $1$, rendering the computational overhead of global attention negligible. Therefore, the hybrid pattern is exclusively applied to the prefill stage, seamlessly transitioning back to standard attention during decoding.

\subsection{Algorithm and Implementation Details}
\label{subsec:acceleration_implementation}



The unified forward pass for the CoRe-guided acceleration is formalized in Algorithm~\ref{alg:hybrid_attention}.

\begin{algorithm}[ht]
\caption{CoRe-Guided Head-level Hybrid Attention (Forward Pass)}
\label{alg:hybrid_attention}
\textbf{Input}: Hidden states $X$, Position Embeddings $(cos, sin)$ \\
\textbf{Input}: Current layer index $l$, Top-$k$ dense head set $\mathcal{H}_{\text{dense}}$ \\
\textbf{Parameters}: $\texttt{FULL\_ATTN\_FLAG} = 0$, $\texttt{SPARSE\_ATTN\_FLAG} = -1$
\begin{algorithmic}[1]
\STATE $Q, K, V \leftarrow \text{ProjectAndApplyRoPE}(X, cos, sin)$
\STATE $B, H, N, D \leftarrow Q.\text{shape}$ \COMMENT{Batch size, Num heads, Seq length, Head dim}
\STATE $\text{is\_prefill} \leftarrow N > 1$
\IF{\textbf{not} $\text{is\_prefill}$}
    \STATE \COMMENT{Decoding Stage: Use standard FlashAttention for efficiency}
    \STATE $O, \text{weights} \leftarrow \text{StandardAttention}(Q, K, V)$
\ELSE
    \STATE \COMMENT{Prefill Stage: Apply Head-level Hybrid Attention}
    \STATE Initialize $\texttt{head\_mask} \in \mathbb{R}^{H}$ with $\texttt{SPARSE\_ATTN\_FLAG}$
    \STATE Extract dense heads for current layer: $\mathcal{H}_{l} \leftarrow \{h \mid (l, h) \in \mathcal{H}_{\text{dense}}\}$
    \IF{$\mathcal{H}_{l} \neq \emptyset$}
        \STATE $\texttt{head\_mask}[\mathcal{H}_{l}] \leftarrow \texttt{FULL\_ATTN\_FLAG}$ \COMMENT{Elevate target heads to dense}
    \ENDIF
    \STATE Define $\texttt{streaming\_info} \leftarrow [4, 32] \times H$ \COMMENT{Set attention sinks and window size}
    \STATE Flatten $Q, K, V$ to shape $(B \times N, H, D)$ for var-len kernel
    \STATE $O \leftarrow \texttt{block\_streaming\_attn\_func}($
    \STATE \quad $Q, K, V, \texttt{head\_mask}, \texttt{streaming\_info}, \text{is\_causal}=\text{True}$
    \STATE $)$
    \STATE Reshape $O$ back to $(B, N, H, D)$
\ENDIF
\STATE $O \leftarrow \text{OutputProjection}(O)$
\RETURN $O, \text{None}$
\end{algorithmic}
\end{algorithm}

By utilizing this fused kernel strategy, we effectively convert the theoretical $O(N^2)$ complexity of standard attention into $O(N \cdot w)$ for the vast majority of heads. As empirically demonstrated in our main experiments (Table~\ref{tab:model_performance_updated}), this system-level optimization yields substantial prefill speedups (up to $2.3\times$) while simultaneously preserving, and in some granular reasoning tasks even slightly enhancing, the multimodal comprehension capabilities of the baseline models.

\subsection{Long-Video Inference Performance of the Video-MME Benchmark}
\label{subsec:videomme}

To evaluate the acceleration strategy on long-context multimodal tasks, we benchmarked the models on the Video-MME dataset. This benchmark tests model performance across perception, recognition, problem-solving, and reasoning. As shown in Table~\ref{tab:videomme_results}, our sparse attention configurations match or occasionally exceed the performance of fully dense baselines.

Overall performance remains stable across all three model families under high sparsity. Instead of degrading, models frequently show slight improvements at specific sparsity levels. For LLaVA-OneVision, the dense baseline scores an average of 56.3. When retaining only 19.1\% and 25.5\% of the attention heads, this average increases to 56.5 and 56.6. Similarly, Qwen3-VL-8B matches its dense baseline average of 65.0 while using only 2.6\% of its attention heads. This performance stabilization suggests an implicit regularization effect. Dense attention mechanisms in long-video contexts tend to aggregate irrelevant background noise. Masking non-essential heads forces the model to rely on specialized CoRe heads, filtering out visual distractions and focusing computation on relevant spatiotemporal information.

The Qwen3-VL family results demonstrate a scaling pattern for attention head redundancy. The 32B model maintains stable performance under extreme sparsity. Retaining 4.9\% of its heads yields a score of 69.7, a 0.2-point decrease from the dense baseline of 69.9. At a 1.2\% retention rate, performance slightly decreases to 69.4. This scaling behavior indicates that visual information routing becomes more concentrated as model capacity increases. Larger models exhibit higher functional redundancy in their attention layers, permitting aggressive pruning during prefill without affecting reasoning accuracy.

Examining specific Video-MME sub-tasks shows how sparsity affects different cognitive dimensions. Spatial perception and OCR metrics frequently improve under sparse conditions. LLaVA's spatial perception increases from 55.6 in the dense setting to 59.3 in the 19.1\% configuration, and Qwen-32B's OCR reaches 78.4 in the 1.2\% configuration, compared to the 76.3 baseline. This indicates that isolating CoRe heads aids in localizing spatial details and text within noisy frames. Higher-order reasoning tasks, such as action, object, and spatio-temporal reasoning, are largely unaffected by head masking. Qwen-8B maintains a score of 80.4 in spatial reasoning across most sparsity levels, confirming that the retained heads capture the semantic logic necessary for video comprehension.

\input{Table/VideoMME}


\subsection{Prefill Latency Benchmarking}
\label{sec:appendix_benchmark_methodology}

We designed a controlled micro-benchmarking protocol to evaluate the speedup achieved by our CoRe-guided hybrid attention mechanism. This protocol isolates the self-attention bottleneck within the language modeling backbone, allowing us to precisely measure the prefill latency across varying context windows ranging from 8k to 128k tokens. 

In Multimodal Large Language Models (MLLMs) like Qwen3-VL, the $O(N^2)$ computational complexity is primarily localized within the self-attention layers of the core language model, rather than the vision encoder. To isolate this scaling behavior and eliminate the constant-time overhead of visual feature extraction, our benchmarking script directly targets the text-conditional generation module. We simulate multimodal long-context inputs by generating randomized discrete token sequences and fully unmasked tensors of length $N$. This setup ensures that the measured latency strictly reflects the computational cost of the attention operations.

Prior to the forward pass, we leverage the empirical Retrieval Attention Mass (RAM) scores to construct a static \texttt{block\_list}. This list dictates the specific layers and heads designated for dense full attention versus Stream Sparse Attention. We inject this configuration into our Triton-based attention kernel, ensuring that the hardware dispatch aligns perfectly with our theoretical attention distribution.

To guarantee precise timing, we employ CUDA event tracking, which circumvents CPU-GPU asynchronous execution discrepancies and Python interpreter overhead. Before recording the measurements, we execute two warmup forward passes to initialize the CUDA context, allocate KV-cache buffers, and stabilize GPU clock speeds. Subsequently, we perform five independent forward passes for each sequence length. We enforce strict synchronization before and after each pass to capture the exact physical GPU execution time. The final prefill latency is reported as the arithmetic mean of these five runs.

Additionally, to evaluate memory scalability, the script is designed to catch \texttt{OutOfMemoryError} exceptions. If a specific context length exceeds the VRAM capacity, the process logs the failure, clears the CUDA cache, and safely proceeds to the next configuration. This mechanism allows us to identify the exact context-length boundaries where standard dense attention fails but our hybrid attention continues to operate successfully.

\section{Limitation}
\label{app:limitation}
While this study provides robust mechanistic insights into the functional sparsity of Multimodal Large Language Models (MLLMs), several limitations warrant future investigation. Primarily, the formulation of our core metric, Retrieval Attention Mass (RAM), relies heavily on explicit ground-truth spatial or temporal annotations (e.g., bounding boxes or video tubes) to locate target visual entities, which complicates the identification of CoRe heads in entirely unannotated or purely abstract reasoning contexts. Furthermore, although our evaluations span diverse, representative late-fusion transformer architectures, it remains an open question whether this distinct functional dichotomy universally emerges in natively multimodal early-fusion models or non-transformer frameworks. Finally, our proposed CoRe-guided hybrid attention mechanism currently employs a deterministic, static head configuration during the prefill stage; while highly efficient on standard benchmarks, advancing towards a dynamic, input-adaptive routing strategy could further enhance model robustness for complex, out-of-distribution queries.

%% file: Table/model.tex
\begin{table}[htbp]
\centering
\small
\caption{Overview of Evaluated MLLMs and their Visual Processing Paradigms}
\label{tab:model_configurations}
\begin{tabularx}{\textwidth}{@{} c c c X @{}}
\toprule
\textbf{Model Family} & \textbf{Variants} & \textbf{Bridging Mechanism} & \textbf{Visual Processing Strategy} \\ 
\midrule
\textbf{Qwen3-VL~\citep{Qwen3-VL}} & 4B, 8B, 32B & Vision-Language Adapter & Local pooling ($2 \times 2$ downsampling) resulting in $c=4$ sequence compression. \\ 
\addlinespace
\textbf{LLaVA-OneVision~\citep{li2025llavaonevision}} & 7B & MLP Projector & Deterministic expansion into fixed-length, continuous 1D token chunks. \\ 
\addlinespace
\textbf{InternVL3.5~\citep{wang2025internvl3_5}} & 8B & Dynamic Cumulation & Adaptive image partitioning based on native aspect ratios (1 to 12 patches). \\ 
\bottomrule
\end{tabularx}
\end{table}

%% file: Table/VideoMME.tex
\definecolor{baselinecolor}{gray}{0.95}
\begin{table*}[t]
\centering
\renewcommand{\arraystretch}{1.2}
\setlength{\tabcolsep}{3.5pt} 
\caption{Overall evaluation results on the VideoMME benchmark. Best results per model family are \textbf{bolded}. The subscript indicates the absolute difference compared to the corresponding Dense baseline.}
\label{tab:videomme_results}
\resizebox{\textwidth}{!}{
\begin{tabular}{l | r | ccc cc cc cccc c | c}
\toprule
\multirow{2.5}{*}{\textbf{Model}} & \multirow{2.5}{*}{\textbf{Config}} & \multicolumn{3}{c}{\textbf{Perception}} & \multicolumn{2}{c}{\textbf{Recognition}} & \multicolumn{2}{c}{\textbf{Problem}} & \multicolumn{4}{c}{\textbf{Reasoning}} & \multirow{2.5}{*}{\textbf{Synopsis}} & \multirow{2.5}{*}{\textbf{Avg.}} \\
\cmidrule(lr){3-5} \cmidrule(lr){6-7} \cmidrule(lr){8-9} \cmidrule(lr){10-13}
& & Temp. & Spat. & Attr. & Act. & Obj. & OCR & Cnt. & Temp. & Spat. & Act. & Obj. & & \\
\midrule

\rowcolor{baselinecolor}
\cellcolor{white} & Dense & 58.2 & 55.6 & \textbf{69.4} & 53.7 & \textbf{62.7} & \textbf{58.3} & 35.8 & 37.3 & 76.8 & 51.2 & \textbf{55.3} & 71.2 & 56.3 \\
  & 2.6\% & 54.5\down{3.7} & 57.4\up{1.8} & 62.6\down{6.8} & 51.4\down{2.3} & 56.5\down{6.2} & 52.5\down{5.8} & 38.4\up{2.6} & 36.2\down{1.1} & 75.0\down{1.8} & 51.9\up{0.7} & 54.6\down{0.7} & 67.5\down{3.7} & 54.0\down{2.3} \\
  & 3.8\% & 58.2\same{} & \textbf{59.3}\up{3.7} & 66.7\down{2.7} & 53.0\down{0.7} & 57.9\down{4.8} & 56.1\down{2.2} & 36.9\up{1.1} & 37.3\same{} & 73.2\down{3.6} & 51.6\up{0.4} & 55.1\down{0.2} & 69.7\down{1.5} & 55.1\down{1.2} \\
  & 6.4\% & 60.0\up{1.8} & 57.4\up{1.8} & 67.1\down{2.3} & 53.0\down{0.7} & 58.8\down{3.9} & 57.6\down{0.7} & 38.1\up{2.3} & 36.2\down{1.1} & 71.4\down{5.4} & 50.9\down{0.3} & 54.4\down{0.9} & 70.0\down{1.2} & 55.2\down{1.1} \\
  & 12.8\% & 60.0\up{1.8} & 53.7\down{1.9} & 66.7\down{2.7} & 54.0\up{0.3} & 56.8\down{5.9} & 56.1\down{2.2} & 38.1\up{2.3} & \textbf{37.9}\up{0.6} & 80.4\up{3.6} & 50.9\down{0.3} & 52.6\down{2.7} & 70.6\down{0.6} & 55.0\down{1.3} \\
  & 19.1\% & \textbf{61.8}\up{3.6} & \textbf{59.3}\up{3.7} & 68.0\down{1.4} & 53.0\down{0.7} & 61.0\down{1.7} & 56.8\down{1.5} & 39.2\up{3.4} & 37.3\same{} & \textbf{82.1}\up{5.3} & 51.2\same{} & \textbf{55.3}\same{} & \textbf{72.1}\up{0.9} & 56.5\up{0.2} \\
\multirow{-7.5}{*}{LLaVA-OneVis.} & 25.5\% & 60.0\up{1.8} & 55.6\same{} & 66.2\down{3.2} & \textbf{54.3}\up{0.6} & 60.7\down{2.0} & 56.1\down{2.2} & \textbf{40.7}\up{4.9} & \textbf{37.9}\up{0.6} & \textbf{82.1}\up{5.3} & \textbf{52.3}\up{1.1} & \textbf{55.3}\same{} & 71.8\up{0.6} & \textbf{56.6}\up{0.3} \\
\midrule

\rowcolor{baselinecolor}
\cellcolor{white} & Dense & \textbf{78.2} & 66.7 & \textbf{76.1} & \textbf{65.5} & 68.4 & 66.2 & 45.9 & \textbf{52.5} & \textbf{80.4} & 60.0 & \textbf{61.9} & 78.6 & \textbf{65.0} \\
  & 1.7\% & 76.4\down{1.8} & 66.7\same{} & 74.8\down{1.3} & 63.9\down{1.6} & 68.4\same{} & 66.9\up{0.7} & \textbf{47.4}\up{1.5} & 50.3\down{2.2} & \textbf{80.4}\same{} & 59.6\down{0.4} & 61.7\down{0.2} & 79.3\up{0.7} & 64.7\down{0.3} \\
  & 2.6\% & 74.5\down{3.7} & 66.7\same{} & 75.7\down{0.4} & 64.2\down{1.3} & 68.4\same{} & 67.6\up{1.4} & 46.6\up{0.7} & 51.4\down{1.1} & \textbf{80.4}\same{} & \textbf{60.7}\up{0.7} & \textbf{61.9}\same{} & \textbf{79.9}\up{1.3} & \textbf{65.0}\same{} \\
  & 4.3\% & \textbf{78.2}\same{} & 66.7\same{} & 75.7\down{0.4} & 63.6\down{1.9} & \textbf{68.9}\up{0.5} & 66.2\same{} & 46.6\up{0.7} & 52.0\down{0.5} & \textbf{80.4}\same{} & 60.4\up{0.4} & 61.7\down{0.2} & 78.9\up{0.3} & 64.9\down{0.1} \\
  & 8.7\% & \textbf{78.2}\same{} & 66.7\same{} & 75.2\down{0.9} & 64.5\down{1.0} & 68.1\down{0.3} & 67.6\up{1.4} & 47.0\up{1.1} & 52.0\down{0.5} & \textbf{80.4}\same{} & 60.0\same{} & 61.7\down{0.2} & 79.3\up{0.7} & 64.9\down{0.1} \\
  & 13.0\% & \textbf{78.2}\same{} & \textbf{68.5}\up{1.8} & 74.8\down{1.3} & 64.9\down{0.6} & 68.6\up{0.2} & \textbf{68.3}\up{2.1} & 47.0\up{1.1} & 51.4\down{1.1} & \textbf{80.4}\same{} & 60.4\up{0.4} & 61.7\down{0.2} & 78.9\up{0.3} & \textbf{65.0}\same{} \\
\multirow{-7.5}{*}{Qwen3-VL-8B} & 17.4\% & \textbf{78.2}\same{} & 66.7\same{} & 75.7\down{0.4} & 64.5\down{1.0} & 68.1\down{0.3} & 67.6\up{1.4} & 46.3\up{0.4} & 52.0\down{0.5} & \textbf{80.4}\same{} & \textbf{60.7}\up{0.7} & 61.7\down{0.2} & 78.9\up{0.3} & 64.9\down{0.1} \\
\midrule

\rowcolor{baselinecolor}
\cellcolor{white} & Dense & 81.8 & \textbf{75.9} & \textbf{81.5} & 68.7 & \textbf{72.0} & 76.3 & \textbf{49.6} & \textbf{57.6} & \textbf{87.5} & \textbf{62.5} & 68.5 & 84.2 & \textbf{69.9} \\
  & 1.2\% & \textbf{83.6}\up{1.8} & \textbf{75.9}\same{} & 80.6\down{0.9} & 67.7\down{1.0} & 71.5\down{0.5} & \textbf{78.4}\up{2.1} & 47.8\down{1.8} & 56.5\down{1.1} & 85.7\down{1.8} & 61.1\down{1.4} & \textbf{68.9}\up{0.4} & 83.6\down{0.6} & 69.4\down{0.5} \\
  & 2.4\% & 81.8\same{} & \textbf{75.9}\same{} & 80.2\down{1.3} & 68.4\down{0.3} & 71.8\down{0.2} & 77.7\up{1.4} & 46.6\down{3.0} & 55.4\down{2.2} & 85.7\down{1.8} & 60.7\down{1.8} & 68.5\same{} & 84.2\same{} & 69.1\down{0.8} \\
  & 3.7\% & 81.8\same{} & \textbf{75.9}\same{} & 80.2\down{1.3} & 68.4\down{0.3} & 71.8\down{0.2} & 77.7\up{1.4} & 48.1\down{1.5} & 55.9\down{1.7} & 85.7\down{1.8} & 60.7\down{1.8} & 67.8\down{0.7} & 84.2\same{} & 69.2\down{0.7} \\
  & 4.9\% & \textbf{83.6}\up{1.8} & \textbf{75.9}\same{} & \textbf{81.5}\same{} & \textbf{69.0}\up{0.3} & \textbf{72.0}\same{} & 77.0\up{0.7} & 48.1\down{1.5} & 56.5\down{1.1} & \textbf{87.5}\same{} & 60.0\down{2.5} & 68.5\same{} & \textbf{85.1}\up{0.9} & 69.7\down{0.2} \\
  & 6.1\% & 81.8\same{} & \textbf{75.9}\same{} & 80.6\down{0.9} & 68.7\same{} & \textbf{72.0}\same{} & 77.0\up{0.7} & 47.8\down{1.8} & 57.1\down{0.5} & \textbf{87.5}\same{} & 60.7\down{1.8} & 68.7\up{0.2} & \textbf{85.1}\up{0.9} & 69.6\down{0.3} \\
\multirow{-7.5}{*}{Qwen3-VL-32B} & 7.3\% & 81.8\same{} & \textbf{75.9}\same{} & 81.1\down{0.4} & 68.4\down{0.3} & \textbf{72.0}\same{} & 77.0\up{0.7} & 48.5\down{1.1} & 55.9\down{1.7} & \textbf{87.5}\same{} & 61.4\down{1.1} & 68.7\up{0.2} & 84.5\up{0.3} & 69.6\down{0.3} \\
\bottomrule
\end{tabular}
}
\end{table*}